\documentclass[10pt,twocolumn,letterpaper]{article}

\usepackage{abstract}
\usepackage{cvpr}
\usepackage{times}
\usepackage{epsfig}
\usepackage{graphicx}
\usepackage{amsmath}
\usepackage{amssymb}
\let\checkmark\undefined  % added for arXiv 
\usepackage{dsfont}
\usepackage{framed}
\usepackage{wrapfig}
\usepackage{cite}
\usepackage{dingbat}
\usepackage{bbding}
\usepackage{balance}
\usepackage{lipsum}
\usepackage[small]{caption}
\usepackage{float}
\usepackage{array}
\usepackage{enumitem}
\usepackage{multirow}
\usepackage{diagbox}
\usepackage{subfigure}
\usepackage{caption}
\usepackage{xcolor}
\usepackage[toc,page]{appendix}

\usepackage{hyperref}  % replace above line for arXiv

\cvprfinalcopy % *** Uncomment this line for the final submission

\def\cvprPaperID{1022} % *** Enter the CVPR Paper ID here

\ifcvprfinal\fi
\begin{document}

\title{\vspace{-5mm}HUMBI: A Large Multiview Dataset of Human Body Expressions}

\author{
\hspace{-2mm}\thanks{Both authors contributed equally to this work}\hspace{1mm}
Zhixuan Yu$^\dagger$
\hspace{15mm}$^\ast$Jae Shin Yoon$^\dagger$
\hspace{15mm}In Kyu Lee$^\dagger$
\hspace{15mm}Prashanth Venkatesh$^\dagger$
\\
\hspace{1mm}Jaesik Park$^\ddagger$
\hspace{25mm}Jihun Yu$^\sharp$
\hspace{25mm}Hyun Soo Park$^\dagger$
\\
\hspace{-5mm}$^\dagger$University of Minnesota
\hspace{20mm}
$^\ddagger$POSTECH
\hspace{20mm}
$^\sharp$BinaryVR
\\
{\tt\small \{yu000064, jsyoon, leex7424, venka220, hspark\}@umn.edu}
\\
{\tt\small jaesik.park@postech.ac.kr, jihun.yu@binaryvr.com}
}
\twocolumn[{%
\maketitle
\thispagestyle{empty}
\vspace{-12mm}
	\begin{center}
		\centering
	\includegraphics[trim=0 90mm 0 48mm, clip, width=\textwidth]{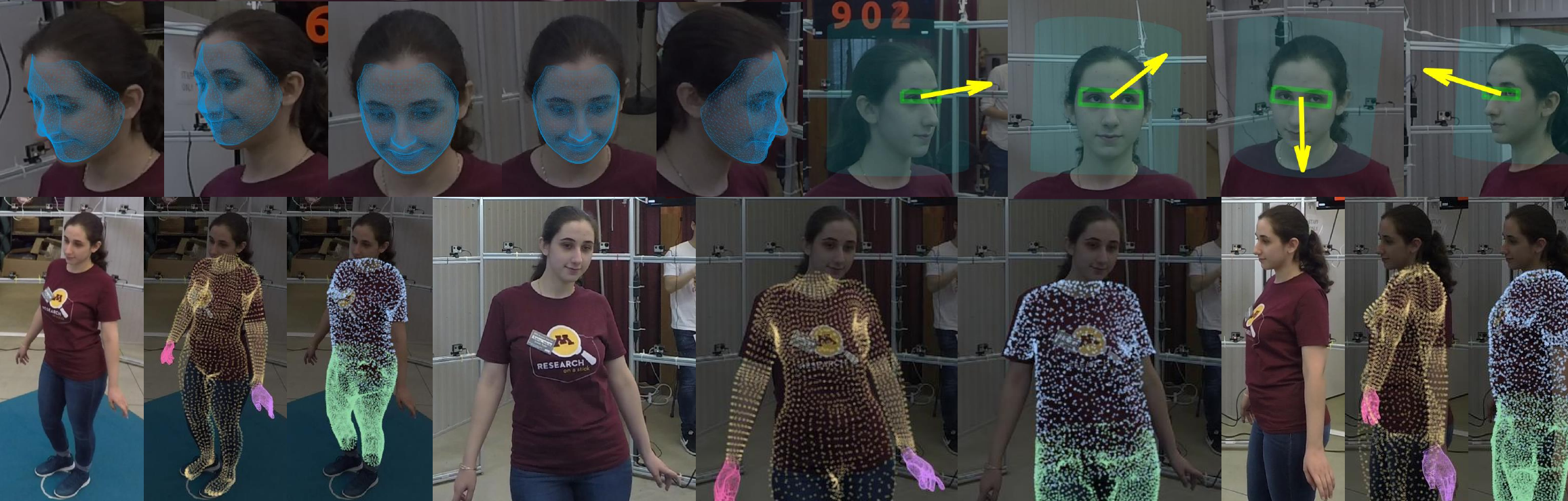}
	\vspace{-5mm}
	\captionof{figure}{We present a new large dataset of multiview human body expressions for modeling view-specific appearance and geometry. 107 synchronized cameras capture the expressions of 772 distinctive subjects. We focus on five elementary expressions: face (blue), gaze (yellow), hand (pink and purple), body (light orange), and garment including top (light blue) and bottom (light green).} 
	\label{fig:teaser_big}
 	\vspace{4mm}
\end{center}	
\label{fig:my_label}
}]
\saythanks

\begin{abstract}
This paper presents a new large multiview dataset called HUMBI for human body expressions with natural clothing. The goal of HUMBI is to facilitate modeling view-specific appearance and geometry of gaze, face, hand, body, and garment from assorted people. 107 synchronized HD cameras are used to capture 772 distinctive subjects across gender, ethnicity, age, and physical condition. With the multiview image streams, we reconstruct high fidelity body expressions using 3D mesh models, which allows representing view-specific appearance using their canonical atlas. We demonstrate that HUMBI is highly effective in learning and reconstructing a complete human model and is complementary to the existing datasets of human body expressions with limited views and subjects such as MPII-Gaze, Multi-PIE, Human3.6M, and Panoptic Studio datasets.  

\end{abstract}

%%%%%%%%% BODY TEXT
\section{Introduction}\label{sec:intro}

We express sincere intent, emotion, and attention through our {\em honest body signals}~\cite{pentland:2008}, including gaze, facial expression, and gestures. Modeling and photorealistic rendering of such body signals are, therefore, the core enabler of authentic telepresence. However, it is challenging due to the complex physical interactions between texture, geometry, illumination, and viewpoint (e.g., translucent skins, tiny wrinkles, and reflective fabric). Recently, pose- and view-specific models by making use of a copious capacity of neural encoding~\cite{LOMBARDI:2018,armando:2018} substantially extend the expressibility of existing linear models~\cite{cootes:2001}. So far, these models have been constructed by a sequence of the detailed scans of a target subject using dedicated camera infrastructure (e.g., multi-camera systems~\cite{joo_cvpr_2014,Bee10,Wenger:2005}). Looking ahead, we would expect a new versatile model that is applicable to the general appearance of assorted people without requiring the massive scans for every target subject.

\begin{figure}[t]
    \begin{center}
        \includegraphics[width=0.47\textwidth]{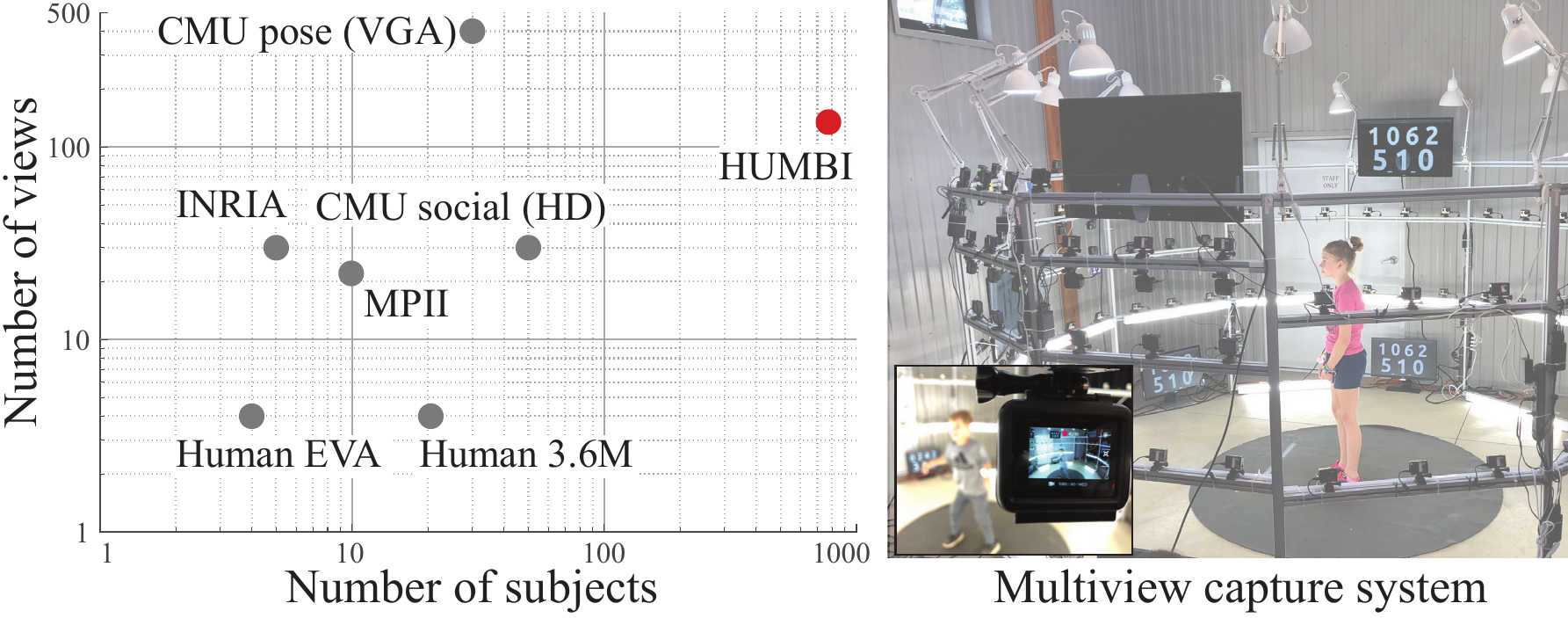}
    \end{center}
    \vspace{-5mm}
    \caption{We present HUMBI that pushes towards two extremes: views and subjects. The view-specific appearance measured by 107 HD cameras regarding five elementary body expressions for 772 distinctive subjects.}
    \label{fig:position}
    \vspace{-4mm}
\end{figure}

Among many factors, what are the core resources to build such a generalizable model? We argue that the data that can span an extensive range of appearances from numerous shapes and identities are prerequisites. To validate our conjecture, we present a new dataset of human body expressions called~\textit{HUMBI} (HUman Multiview Behavioral Imaging) that pushes to two extremes: views and subjects. As of Nov 2019\footnote{In a contract with public event venues, the dataset is expected to grow every year.}, the dataset is composed of 772 distinctive subjects with natural clothing across diverse age, gender, ethnicity, and physical condition captured by 107 HD synchronized cameras (68 cameras facing at frontal body). Comparing to existing datasets for human body expressions such as CMU Panoptic Studio~\cite{joo:2015,joo:2019}, MPII~\cite{Dyna:SIGGRAPH:2015,pons2017clothcap}, and INRIA~\cite{Knossow:2008}, HUMBI presents the unprecedented scale visual data (Figure~\ref{fig:position}) that are ideal for learning the detailed appearance and geometry of five elementary human body expressions: gaze, face, hand, body, and garment (Figure~\ref{fig:teaser_big}).

Our analysis shows that HUMBI is effective. We make use of vanilla convolutional neural networks (CNN) to learn view-invariant 3D pose from HUMBI, which quantitatively outperforms the counterpart models trained by existing datasets with limited views and subjects. More importantly, we show that HUMBI is \textit{complementary} to such datasets, i.e., the trained models can be substantially improved by combining with these datasets.

% evaluate view-specific appearance and view-invariant pose, which quantitatively outperforms the counterpart models trained by existing datasets with limited views and subjects. 

% a vanilla auto-encoder, and view-invariant pose detector can be 

% large scale multiview image streams, and view-invariant pose detection can be learned. More importantly, HUMBI is complementary to existing datasets of human expressions with limited views and subjects such as MPII Gaze~\cite{Gross2009}, Human 3.6M~\cite{h36m_pami}, and CMU hand~\cite{simon2017hand}, i.e., by combining these datasets, the performance of the trained detectors is substantially improved without sophisticated network tuning.  

% we show that its complementary property to existing datasets such as MPII Gaze~\cite{Gross2009}, Human 3.6M~\cite{h36m_pami}, and CMU hand~\cite{simon2017hand}. The recognition performance can be significantly improved as HUMBI enforces view-invariance without sophisticated network tuning. 

The main properties of HUMBI are summarized below. 
(1) Complete: it captures the total body, including gaze, face, hand, foot, body, and garment to represent holistic body signals~\cite{Joo:2018}, e.g., perceptual asynchrony between the face and hand movements. 
(2) Dense: 107 HD cameras create a dense light field that observe the minute body expressions with minimal self-occlusion. This dense light field allows us to model precise appearance as a function of view~\cite{LOMBARDI:2018}. 
(3) Natural: the subjects are all voluntary participants (no actor/actress/student/researcher). Their activities are loosely guided by performance instructions, which generates natural body expressions. (4) Diverse: 772 distinctive subjects with diverse clothing styles, skin colors, time-varying geometry of gaze/face/body/hand, and range of motion. (5) Fine: with multiview HD cameras, we reconstruct the high fidelity 3D model using 3D meshes, which allows representing view-specific appearance in its canonical atlas.

% each HD camera measures fine-grained behavioral signals running at 60Hz that can measure the bod where high fidelity recovery is possible.

% To our best knowledge, this is the first dataset that includes a diverse population measured by a large number of the camera array. The core contributions include: (a) The unique design of the portable multi-camera system that can be deployed at public events to capture diverse subjects. (b) The tera-scale multiview visual data that measure the detailed human behaviors at the millimeter scale. (c) The high fidelity appearance models for gaze, face, hands, body, and cloth, and their reconstruction algorithm. (d) Geometrically consistent multiview image annotations via the 2D projection of 3D models. 

% The HUMBI will open up a new opportunity to develop computer vision systems that can precisely decode the minute details of our behaviors in daily living space, which will genuinely facilitate smart home/hospital that assists millions of the elderly, children, and patients. 

\begin{table*}[h]
\vspace{-6mm}
\centering
\scriptsize
\resizebox{0.995\linewidth}{!}{
\begin{tabular}{l|l|l|l|l|l|l|l}
\hline
Dataset & \# of subjects & Measurement method & Gaze & Face & Hand & Body & Cloth\\
\hline
\hline
Columbia Gaze~\cite{smith:2013} & 56 & 5 cameras & \checkmark (fixed) &  & &  & \\
UT-Multiview~\cite{sugano:2014} & 50 & 8 cameras & \checkmark (fixed) &  & &  & \\
Eyediap~\cite{Mora:2014} & 16 & 1 depth camera and 1 HD camera & \checkmark (free) &  & &  & \\
MPII-Gaze~\cite{zhang15_cvpr} & 15 & 1 camera & \checkmark (free)  &  & &  & \\
RT-GENE~\cite{fischer2018rt} & 17 & eyetracking device & \checkmark (free)  &  & &  & \\
\hline
CMU Multi-PIE~\cite{Gross2009} & 337 & 15 cameras &  & \checkmark & &  & \\
3DMM~\cite{blanz:2003} & 200 & 3D scanner &  & \checkmark & &  & \\
BFM~\cite{bfm09} & 200 & 3D scanner &  & \checkmark & &  & \\
ICL~\cite{booth2018large} & 10,000 & 3D scanner &   & \checkmark & &  & \\
\hline
NYU Hand~\cite{tompson:2014}& 2 (81K samples) & Depth camera &   &  & \checkmark &  & \\
HandNet~\cite{Wetzler:2016}& 10 (213K samples) & Depth camera and magnetic sensor &   &  & \checkmark &  & \\
BigHand 2.2M~\cite{yuan:2017}& 10 (2.2M samples) & Depth camera and magnetic sensor &   &  & \checkmark &  & \\
RHD~\cite{zimmermann2017learning}& 20 (44K samples) &  N/A (synthesized) &   &  & \checkmark &  & \\
STB~\cite{zhang2017hand}& 1 (18K samples) &  1 pair of stereo cameras &   &  & \checkmark &  & \\
FreiHand~\cite{Freihand2019}& N/A (33K samples) & 8 cameras &   &  & \checkmark &  & \\
\hline
CMU Mocap & $\sim$100& Marker-based & &  & & \checkmark & \\ 
CMU Skin Mocap~\cite{park:2006} & $<$10 & Marker-based & & \checkmark & & \checkmark & \\
INRIA~\cite{Knossow:2008} & N/A & Markerless (34 cameras) & & & & \checkmark & \checkmark (natural)\\
Human EVA~\cite{sigal2010humaneva} & 4 & Marker-based and Markerless (4-7 cameras) & & & & \checkmark & \\
Human 3.6M~\cite{h36m_pami} & 11 & Markerless (depth camera and 4 HD cameras) & & & & \checkmark & \\
Panoptic Studio~\cite{Joo_2017_TPAMI,simon2017hand} & $\sim$100 & Markerless (31 HD and 480 VGA cameras) &  &  & \checkmark & \checkmark  & \\
Dyna~\cite{Dyna:SIGGRAPH:2015} & 10 & Markerless (22 pairs of stereo cameras) &  &  & &  \checkmark & \\
ClothCap~\cite{pons2017clothcap} & 10 & Markerless (22 pairs of stereo cameras) &  &  &  & & \checkmark (synthesized)\\
BUFF~\cite{Zhang_2017_CVPR} & 5 & Markerless (22 pairs of stereo cameras) &  &  &  & \checkmark & \checkmark (natural) \\
3DPW~\cite{vonMarcard2018} & 7 & Marker-based (17 IMUs) and Markerless (1 camera + 3D scanner) &  &  &  & \checkmark & \checkmark (natural) \\
% 3DPW~\cite{vonMarcard2018} & 7 & Marker-based and Markerless (1 camera + 3D scanner) &  &  &  & \checkmark & \checkmark (natural) \\
TNT15~\cite{vonPon2016a} & 4 & Marker-based (10 IMUs) and Markerless (8 cameras + 3D scanner)  &  &  &  & \checkmark & \\
% TNT15~\cite{vonPon2016a} & 4 & Marker-based and Markerless (8 cameras + 3D scanner)  &  &  &  & \checkmark & \\
D-FAUST\cite{dfaust:CVPR:2017} & 10 & Markerless (22 pairs of stereo cameras) &  &  &  & \checkmark & \\
\hline
\hline
HUMBI & 772  & Markerless (107 HD cameras) & \checkmark (free) & \checkmark & \checkmark & \checkmark  & \checkmark (natural)\\
\hline
\end{tabular}}
\vspace{-3mm}
\caption{Human body expression datasets.}
\vspace{-3mm}
\label{table:dataset}
\end{table*}

\section{Related Work}\label{sec:related}

% Humans transmit and respond to many different behavioral signals such as gaze movement, facial expression, and body gestures when they interact with others~\cite{pentland:2008,vinciarelli:2009}. Effective signaling and interpretation of signals are the basis of successful social performance, for example, in business~\cite{pentland:2004,curhan:2007,carney:2010}. Researchers have developed various computational models to measure, model, and predict the behavioral signals~\cite{pentland:2005,eagle:2006}. Some behavioral signals such as hand-flapping, repeating sounds, and deficits of joint attention have shown to be early markers of the autistic spectrum disorder, and computational tools have been designed to detect these symptoms~\cite{rehg:2013,ousley:2012}. The data of behavioral signals is the key enabling factor, which builds computational models. Here, 

We briefly review the existing datasets for modeling human body expressions: gaze, face, hand, body, and garment. These datasets are summarized in Table~\ref{table:dataset}.

\noindent\textbf{Gaze} Columbia Gaze dataset~\cite{smith:2013} and UT-Multiview dataset~\cite{sugano:2014} have been captured in a controlled environments where the head poses are fixed. In subsequent work, such constraints have been relaxed. Eyediap dataset~\cite{Mora:2014} captured gaze while allowing head motion, providing natural gaze movements. MPII-Gaze dataset~\cite{zhang15_cvpr} measured in-the-wild gaze from laptops, including 214K images across 15 subjects. This contains a variety of appearance and illumination. RT-GENE dataset~\cite{fischer2018rt} takes a step further by measuring free-ranging point of regard where the ground truth was obtained by using motion capture of mobile eye-tracking glasses.

\noindent\textbf{Face} 3D Morphable Model (3DMM)~\cite{blanz:2003} was constructed by 3D scans of large population to model the complex geometry and appearance of human faces. For instance, 3D faces were reconstructed by leveraging facial landmarks~\cite{jourabloo2016large, sagonas2016300, le2012interactive, sagonas2013semi,belhumeur2013localizing}, and dense face mesh~\cite{tewari17MoFA, feng2018prn}. Notably, 3DMM is fitted to 60K samples from several face alignment datasets~\cite{messer1999xm2vtsdb,sagonas2013300,zhou2013extensive,belhumeur2013localizing, zhu2012face} to create the 300W-LP dataset~\cite{zhu2016face}. For facial appearance, a deep appearance model~\cite{LOMBARDI:2018} introduces view-dependent appearance using a conditional variational autoencoder, which outperforms linear active appearance model~\cite{cootes:2001}.

\noindent\textbf{Hand} Dexterous hand manipulation frequently introduces self-occlusion, which makes building a 3D hand pose dataset challenging. A depth image that provides trivial hand segmentation in conjunction with tracking has been used to establish the ground truth hand pose~\cite{tompson:2014,sun:2015,tang:2014,supancic:2015}. However, such approaches still require intense manual adjustments. This challenge was addressed by making use of graphically generated hands~\cite{mueller2018ganerated,zimmermann2017learning,mueller2017real}, which may introduce a domain gap between real and synthetic data. For real data, an auxiliary input such as magnetic sensors was used to precisely measure the joint angle and recover 3D hand pose using forward kinematics~\cite{Wetzler:2016,yuan:2017}. Notably, a multi-camera system has been used to annotate hands using 3D bootstrapping~\cite{simon2017hand}, which provided the hand annotations for RGB data. FreiHAND\cite{Freihand2019} leveraged MANO\cite{MANO:SIGGRAPHASIA:2017} mesh model to represent dense hand pose.
\noindent\textbf{Body} Markerless motion capture is a viable solution to measure dense human body expression at high resolution. For example, multi-camera systems have been used to capture a diverse set of body poses, e.g., actors and actresses perform a few scripted activities such as drinking, answering cellphone, and sitting~\cite{h36m_pami, sigal2010humaneva}. Natural 3D human behaviors were captured in the midst of the role-playing of social events from a multiview system~\cite{Joo_2017_TPAMI}, while those events inherently involve with a significant occlusion by people or objects that inhibit modeling a complete human body. Further, a 4D scanner~\cite{dfaust:CVPR:2017,Dyna:SIGGRAPH:2015} enabled high resolution body capture to construct a parametric human models, e.g., SMPL~\cite{loper2015smpl}. Notably, image-to-surface correspondences on 50K COCO images~\cite{lin2014microsoft} enabled modeling humans from a single view image~\cite{Lassner:UP:2017}. Further, rendering of human model in images could alleviate annotation efforts~\cite{varol17_surreal}.

\noindent\textbf{Clothes} Previous works have proposed to capture the natural cloth deformation in response to human body movement. Cloth regions were segmented in 3D using multiview reconstruction~\cite{white2007capturing,bradley2008markerless}. To ensure the same topology when segmenting the cloth from 3D reconstruction, the SMPL body model can be used to parametrize cloth motion, which produces physically plausible cloth geometry while preserving wrinkle level details~\cite{pons2017clothcap}.

\noindent\textbf{Our Approach} Unlike existing datasets focusing on each body expressions, HUMBI is designed to span geometry and appearance of total body expressions from a number of distinctive subjects using a dense camera array. Our tera-scale multiview visual data provide a new opportunity to generalize pose- and view-specific appearance.

\begin{figure*}[t]
	\begin{center}
	\vspace{0mm}
		\includegraphics[width=\textwidth]{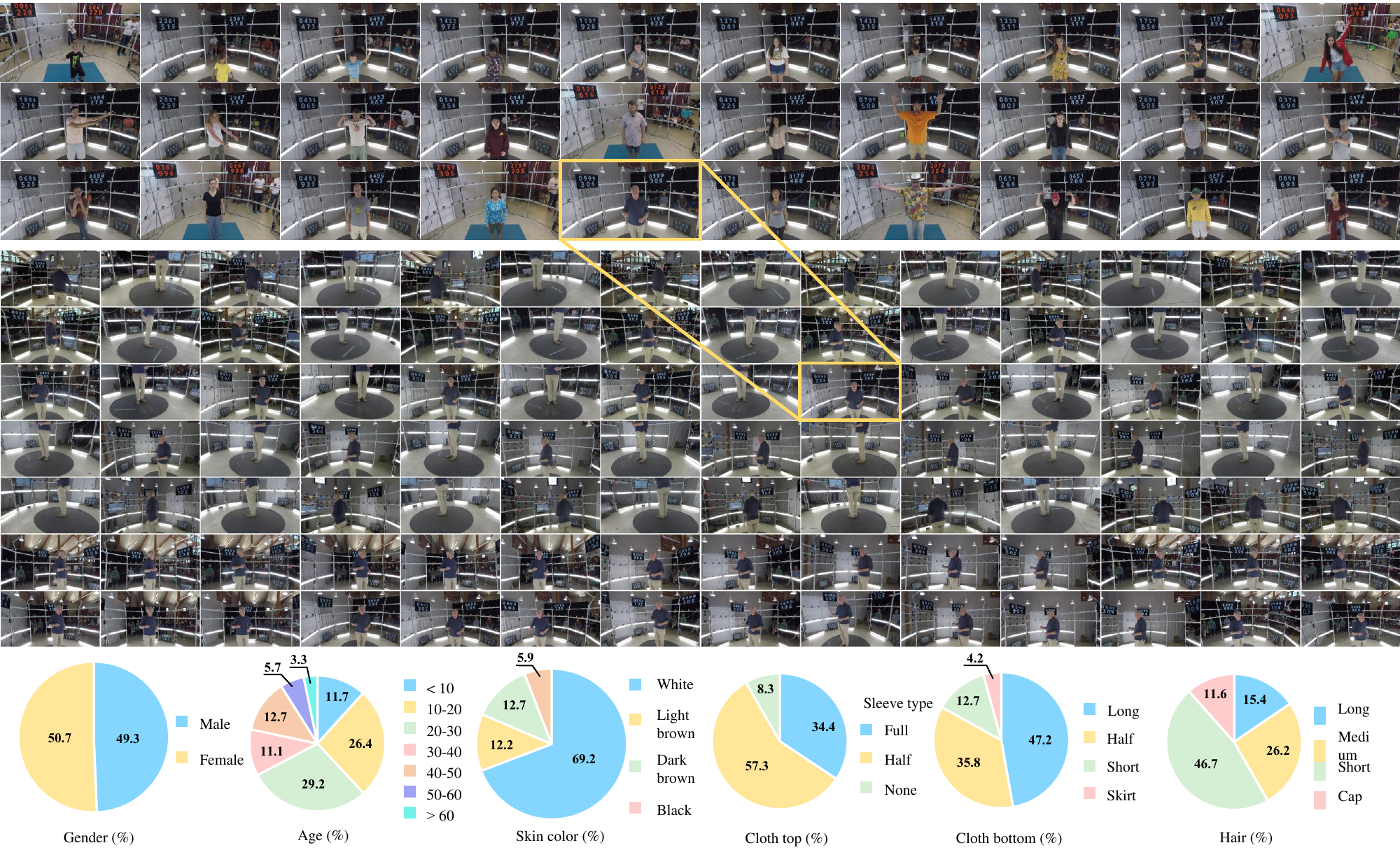}
	\end{center}
	\vspace{-6mm}
    \caption{(Top and bottom) HUMBI includes 772 distinctive subjects across gender, ethnicity, age, clothing style, and physical condition, which generates diverse appearance of human expressions. (Middle) For each subject, 107 HD cameras capture her/his expressions including gaze, face, hand, body, and garment.}
    \label{fig:subject}
    \vspace{-3mm}
\end{figure*}

%\begin{figure}[t]
% 	\begin{center}
% 		\includegraphics[trim={10cm 0cm 0 0},clip,width=3.3in]{./figure/subject1.pdf}
% 	\end{center}
% 	\vspace{-10mm}

%	\begin{center}
% 		\includegraphics[width=3.5in]{./figure/statistic.pdf}
%	\end{center}
%	\vspace{-3mm}
%    \caption{\small \textbf{Top}: Example of subjects. 772 subjects across diverse gender, ethnicity, age were captured by our behavioral imaging system. \textbf{Bottom}: Statistics over 650 subjects.}
%    \label{fig:stat}
%\end{figure}

\section{HUMBI}\label{sec:humbi}

HUMBI is composed of 772 distinctive subjects captured by 107 synchronized HD cameras. 69 cameras are uniformly distributed across dodecagon frame with 2.5m diameter along the two levels of an arc (0.8 m and 1.6 m) where the baseline between adjacent cameras is
approximately 10$^\circ$ (22 cm). Another 38 cameras are distributed across the frontal quadrant of the dodecagon frame (average baseline: 10 cm) to densify cameras used for capturing face/gaze. The dataset includes the five elementary body expressions: gaze, face, hand, body, and garment. We use COLMAP~\cite{schoenberger2016sfm} to calibrate cameras, and upgrade the reconstruction to the metric scale using physical camera baselines. Notable subject statistics includes: evenly distributed gender (50.7\% female; 49.3\% male); a wide range of age groups (11\% of thirties, 29\% of twenties, and 26\% of teenagers); diverse skin colors (black, dark brown, light brown, and white); various styles of clothing (dress, short-/long-sleeve t-shirt, jacket, hat, and short-/long-pants). The statistics are summarized in Figure~\ref{fig:subject}. In this section, we focus on the resulting computational representations while deferring the detailed description of reconstruction approaches to Appendix. 

\noindent\textbf{Notation} We denote our representation of human body expressions as follows:\vspace{-2mm}
\begin{itemize}[leftmargin=*]
    \item[\tiny$\bullet$] Images: $\mathcal{I}=\{\mathbf{I}_i\}$ is a set of multiview images.\vspace{-2mm}
    \item[\tiny$\bullet$] 3D keypoints: $\mathcal{K}$.\vspace{-2mm}
    \item[\tiny$\bullet$] 3D mesh: $\mathcal{M}=\{\mathcal{V}, \mathcal{E}\}$.\vspace{-2mm}
    \item[\tiny$\bullet$] 3D occupancy map: $\mathcal{O}:\mathds{R}^3\rightarrow \{0,1\}$ that takes as input 3D voxel coordinate and outputs binary occupancy.\vspace{-2mm}
    \item[\tiny$\bullet$] Appearance map: $\mathcal{A}:\mathds{R}^2\rightarrow [0,1]^3$ that takes as input atlas coordinate (UV) and outputs normalized RGB values.
\end{itemize}
\vspace{-2mm}

\noindent\textbf{Keypoint} 3D keypoints on face ($\mathcal{K}_{\rm face}$), hands ($\mathcal{K}_{\rm hand}$), and body including feet ($\mathcal{K}_{\rm body}$) are reconstructed by triangulating 2D human keypoint detections~\cite{cao2017realtime} with RANSAC, followed by a nonlinear refinement minimizing geometric reprojection error. When multiple humans are visible, we localize each subject via geometric verification.

\subsection{Gaze}\label{humbi:gaze}
 
HUMBI Gaze contains $\sim$93K images (4 gaze directions $\times\sim$30 views per subject). We represent gaze geometry using a unit 3D vector $\mathbf{g}\in\mathds{S}^2$ with respect to the moving head coordinate system. 

The head coordinate is defined as follows. The origin is the center of eyes, $\mathbf{o}=(\mathbf{p}_{l} + \mathbf{p}_{r})/2$ where $\mathbf{p}_{l}, \mathbf{p}_{r}\in\mathbb{R}^3$ are left and right eye centers. The $x$-axis is the direction along the line joining the two eye centers, $(\mathbf{p}_{l}-\mathbf{o})/\|\mathbf{p}_{l}-\mathbf{o}\|$; the $z$-axis is the direction perpendicular to the plane made of $\mathbf{p}_l$, $\mathbf{p}_{r}$, and $\mathbf{p}_{m}$ where $\mathbf{p}_{m}$ is the center of the mouth, orienting towards the hind face; $y$-axis is defined as a vector orthogonal to both $x$- and $z$-axes under right-hand rule constraint.

\begin{figure*}[t]
\vspace{-4mm}
  \centering
  \subfigure[Gaze appearance]{\label{fig:gaze_uv}\includegraphics[height=0.27\textheight]{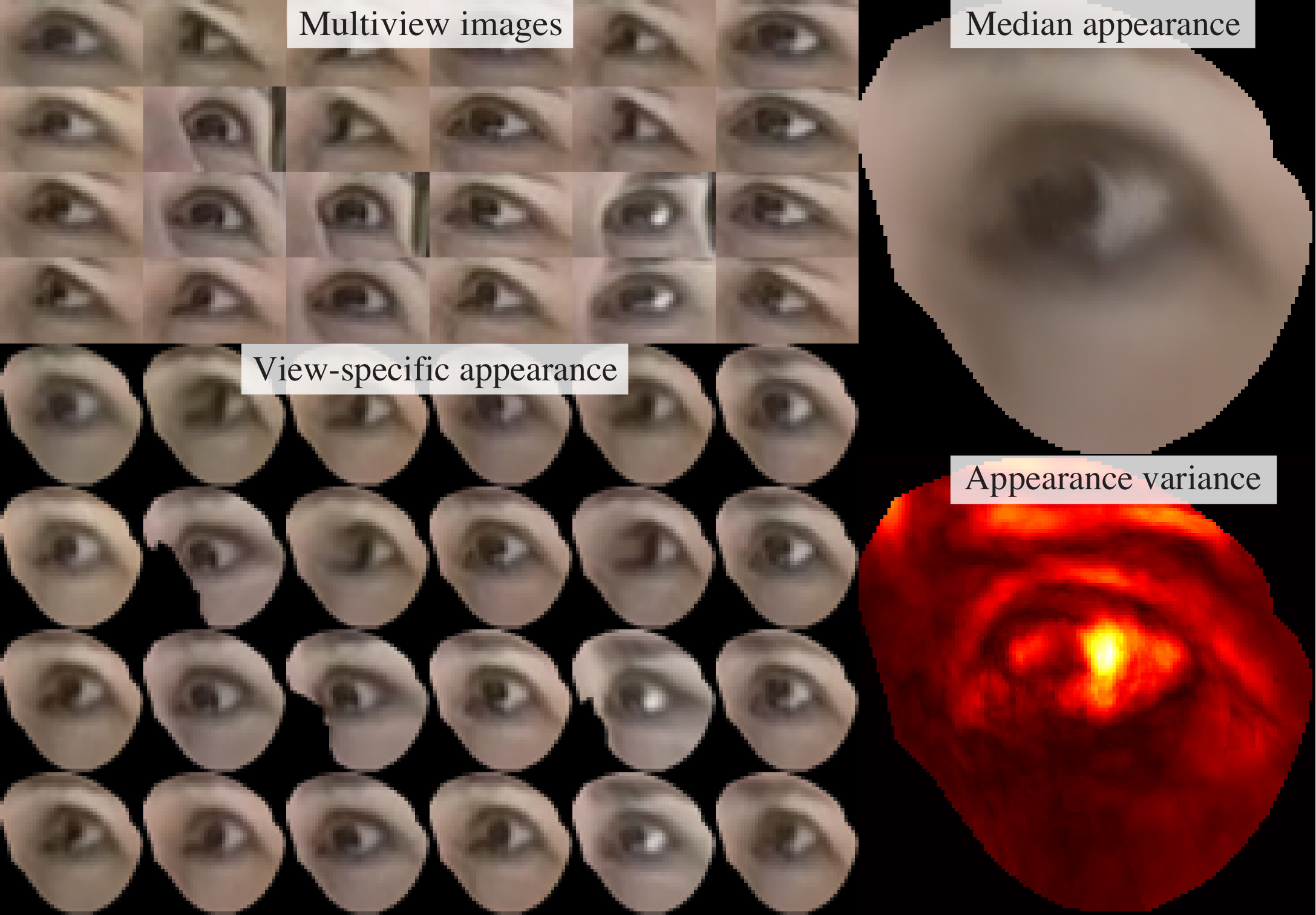}}
  \subfigure[Face appearance]{\label{fig:face_uv}\includegraphics[height=0.27\textheight]{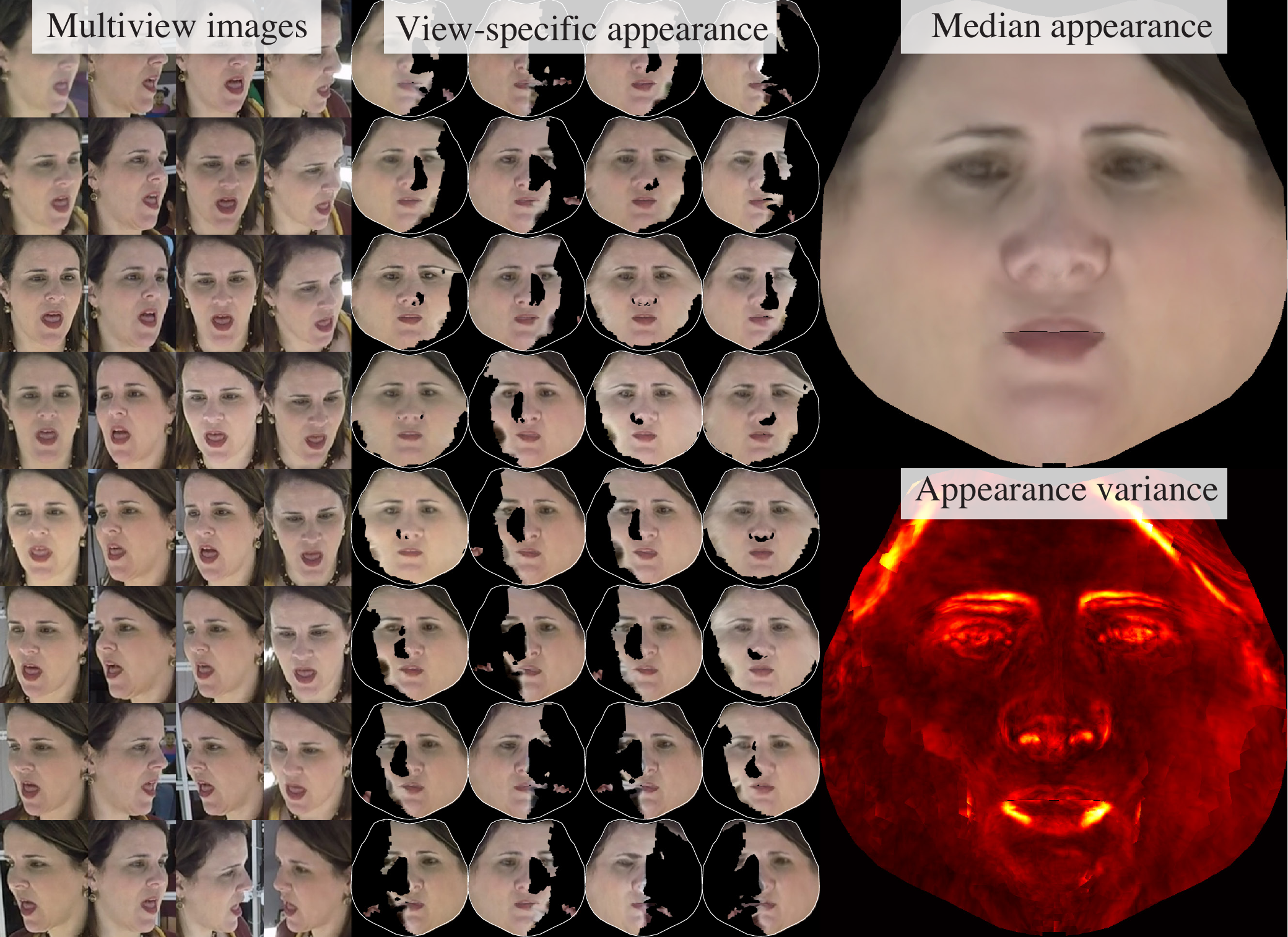}}
  \vskip 0pt  % must add 0pt after \vskip for arXiv
  \vspace{-2mm}
  \subfigure[Hand appearance]{\label{fig:hand_uv}\includegraphics[height=0.247\textheight]{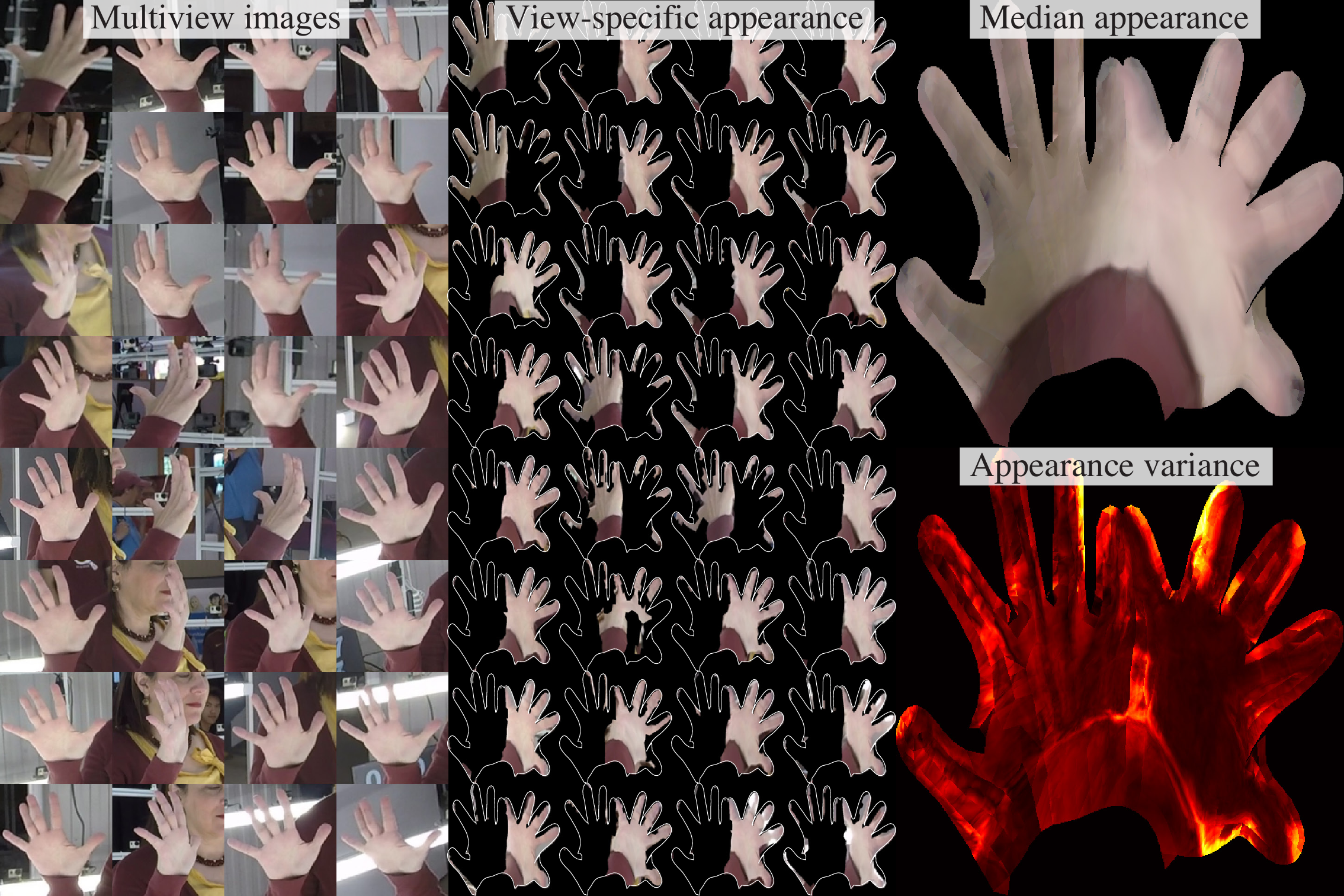}}
  \subfigure[Body appearance]{\label{fig:body_uv}\includegraphics[height=0.247\textheight]{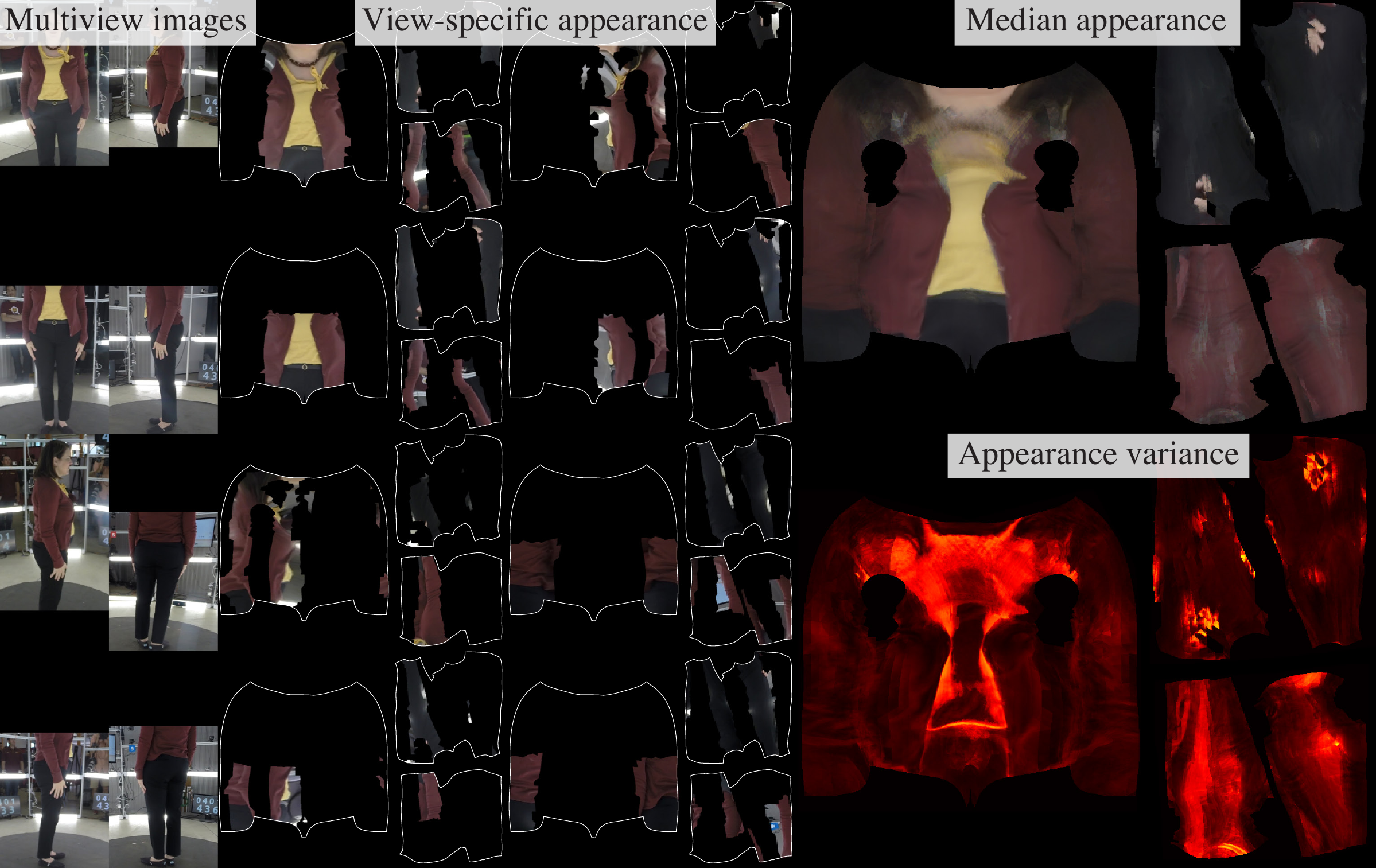}}
  \vspace{-5mm}
  \caption{View-specific appearance rendered from multiview images with median appearance and variance for (a) gaze, (b) face, (c) hand, (d) body.}
  \vspace{-4mm}
\end{figure*}

For eye appearance, we provide two representations: (1) normalized eye patches and (2) pose-independent appearance map. For the normalized eye patches, we warp an eye patch region such that the orientation and distance remain constant across views. RGB values are histogram-equalized. For appearance, we select vertices of eye region in the Surrey face model~\cite{huber2016multiresolution} to build a canonical atlas coordinate (UV) for each eye. We represent view-specific appearance map $\mathcal{A}_{\rm gaze}$ by projecting pixels in the image onto that the atlas coordinate. Figure~\ref{fig:gaze_uv} illustrates view-specific appearance across views with median and variance of appearance. The variance map shows that the appearance is highly dependent on viewpoint in particular in the iris region.

\subsection{Face}\label{humbi:face}

HUMBI Face contains $\sim$17.3M images (330 frames $\times$ 68 views per subject). We represent face geometry using a 3D blend shape model $\mathcal{M}_{face}$ (Surrey~\cite{huber2016multiresolution}) with 3,448 vertices and 6,736 faces. We reconstruct the shape model using 68 facial keypoints ($\mathcal{K}_{\rm face}$) and the associated multiview images ($\mathcal{I}_{\rm face}$), i.e., $\mathcal{M}_{\rm face} = f_{\rm face}(\mathcal{K}_{\rm face}, \mathcal{I}_{\rm face})$ where $f_{\rm face}$ is a face alignment function. We align the face model by minimizing reprojection error over shape, expression, illumination, and texture parameters (see Appendix). Given the reconstructed face mesh model, we construct a view-specific appearance map $\mathcal{A}_{\rm face}$ by projecting pixels in the image onto its canonical atlas coordinate. For each view, the projection map between the image and atlas coordinate is established through the corresponding 3D locations in the reconstructed mesh with bilinear interpolation. Figure~\ref{fig:face_uv} illustrates view-specific appearance across views with median and variance of appearance. The variance map shows that the appearance is dependent on views, e.g. the regions of salient landmarks such as eye, eyebrows, nose, and mouth, which justifies the necessity of view-specific appearance modeling~\cite{LOMBARDI:2018}.

\subsection{Hand}\label{humbi:hand}
HUMBI Hand contains $\sim$24M images (290 frames $\times$ 68 views per subject). We represent hand geometry using a 3D parametric model $\mathcal{M}_{\rm hand}$ (MANO~\cite{MANO:SIGGRAPHASIA:2017}) with 778 vertices and 1,538 faces. We reconstruct the mesh model using hand keypoints ($\mathcal{K}_{\rm hand}$ with 21 keypoints), i.e., $\mathcal{M}_{\rm hand}=f_{\rm hand}(\mathcal{K}_{\rm face})$, where $f_{\rm hand}$ is a hand alignment function. We align the hand model to multiview images by minimizing the Euclidean distance between hand keypoints and the corresponding pose of the mesh model with a $L_2$ parameter regularization. To learn the consistent shape of the hand model for each subject, we infer the maximum likelihood estimate of the shape parameter given the reconstructed keypoints over frames (see Appendix). Given the reconstructed hand mesh model, we construct a view-specific appearance map $\mathcal{A}_{\rm hand}$ by projecting pixels in an image onto the canonical atlas coordinate. Figure~\ref{fig:hand_uv} illustrates view-specific appearance across views with median and variance of appearance. The variance map shows that the appearance is dependent on view points.

% Notice that different from SMPL model, pose parameters of MANO are not axis-angle representation of joint rotations but coefficients of hand pose PCA basis, which allows dimensionality to be reduced and makes fitting to noisy, low-resolution hand data more robust. In practice, we use first 20 pose basis and first shape basis for our data. MANO model is fitted to reconstructed 3D hand keypoints by minimizing: $E^{hand}\big(t,\theta,\beta) = \sum_i^{21} \|\mathbf{K}^h_i - (\mathcal{N}_i(\mathbf{H})+t\big)\|^2 \nonumber+\lambda_\theta\sum_{i=4}^{23}\|{\theta}_i\|^2 +\lambda_\beta\|{\beta}_1\|^2$
% where $\mathbf{K}^h_i$ is $i^{\rm th}$ hand keypoint, and  $\mathcal{N}$ returns a point in shape $\mathbf{H}$ corresponding to the keypoints. $t\in\mathbb{R}^3$ is global translation. $\lambda_\theta$ and $\lambda_\beta$ are weights for pose and shape regularization respectively. For the same subject, initially hand mesh of each frame is reconstructed independently. Then shape parameters are fixed as the median values of all frames. Other parameters are optimized again subsequently. 

% \begin{figure}[t]
% 	\begin{center}
% 		\includegraphics[width=0.47\textwidth]{./figure/body_uv.pdf}
% 	\end{center}
% 	\vspace{-5mm}
%     \caption{\small View-specific body appearance rendered from multiview images with median appearance and variance extracted.}
%     \label{fig:body_uv}
% \end{figure}

\subsection{Body}\label{humbi:body}
Each subject performs a sequence of motion and dance performance, which constitutes $\sim$26M images. Given a set of multiview images at each time instant, we reconstruct a mesh model $\mathcal{M}_{\rm body}$ using body keypoints $\mathcal{K}_{\rm body}$, and occupancy map $\mathcal{O}_{\rm body}$, i.e., $\mathcal{M}_{\rm body} = f_{\rm body}(\mathcal{K}_{\rm body}, \mathcal{O}_{\rm body})$ where $f_{\rm body}$ is a alignment function that matches the surface of $\mathcal{M}_{\rm body}$ to the outer surface of the occupancy map while minimizing the distance between the reconstructed keypoints $\mathcal{K}_{\rm body}$ and the underlying pose of the mesh (see Appendix). We use the SMPL parametric model~\cite{loper2015smpl} that is composed of 4,129 vertices and 7,999 faces without hand and head vertices.

\begin{figure}[t]

	\begin{center}
		\includegraphics[width=3.3in]{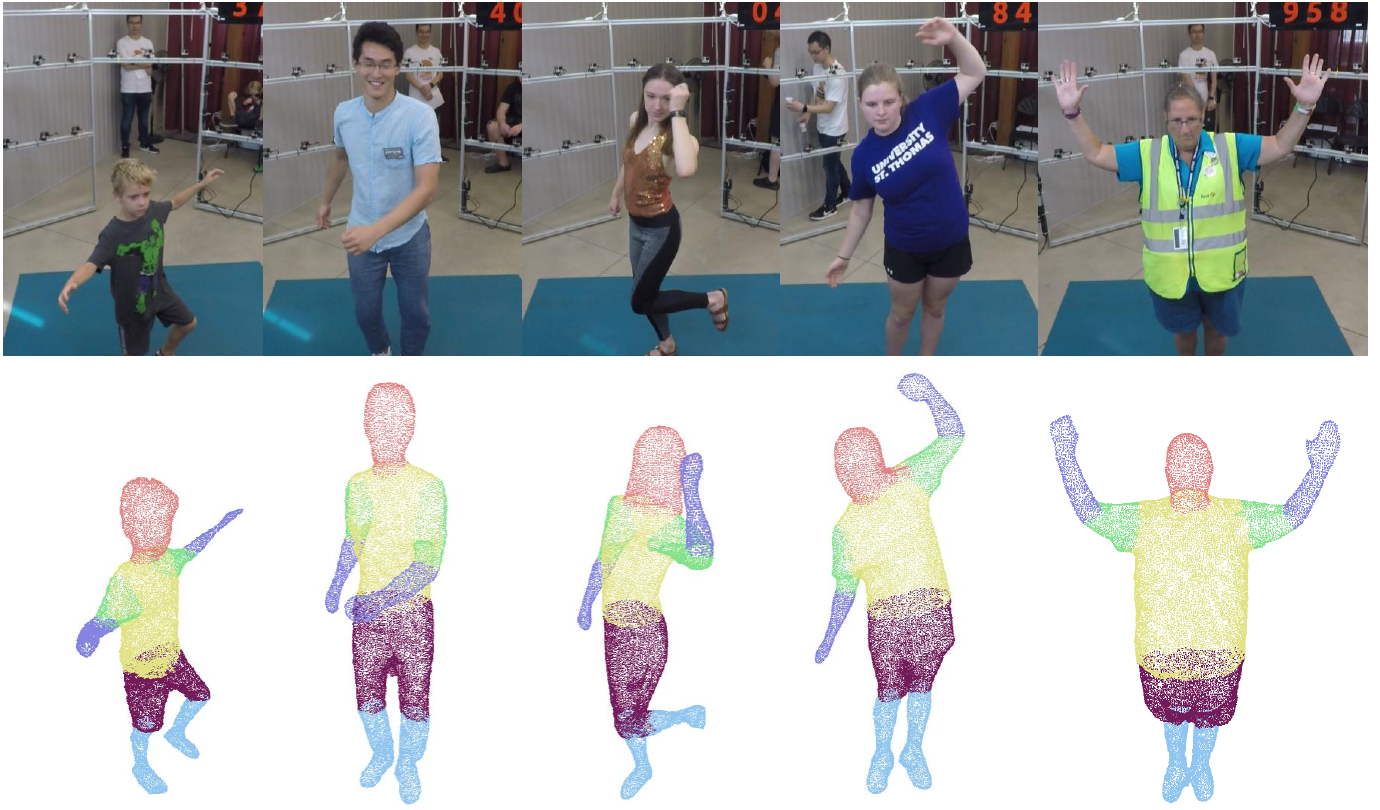}
	\end{center}
	\vspace{-6mm}
    \caption{We reconstruct the body occupancy map and its outer surface using shape-from-silhouette and associate the point cloud with body semantics (head, body, arms, and legs).}
    \label{fig:semantics}
    \vspace{-5mm}
\end{figure}

Shape-from-silhouette\footnote{MultiView stereo~\cite{schoenberger2016sfm} is complementary to the occupancy map.}~\cite{Laurentini94} is used to reconstruct the occupancy map $\mathcal{O}_{\rm body}$. The occupancy map is generated by human body segmentation~\cite{lin2017refinenet}. As a by-product, the semantics (i.e., head, torso, upper arm, lower arm, upper leg, and lower leg) can be labeled at each location in the occupancy map by associating with the projected body label~\cite{yoon20173d} as shown in Figure~\ref{fig:semantics}.  

Given the reconstructed body mesh model, we construct a view-specific appearance map $\mathcal{A}_{\rm body}$ by projecting pixels in an image onto the canonical atlas coordinate. Figure~\ref{fig:body_uv} illustrates view-specific appearance across views with median and variance of appearance. The variance map shows that the appearance is dependent on view points.

\subsection{Garment}\label{humbi:cloth}

Given the body reconstruction, we represent the garment geometry using a garment mesh model $\mathcal{M}_{\rm cloth}$ as similar to~\cite{pons2017clothcap}. An alignment function $\mathcal{M}_{\rm cloth}=f_{\rm cloth}(\mathcal{M}_{\rm body},\mathcal{O}_{\rm body})$ is used to reconstruct the cloth mesh model from the body model and occupancy map. A set of fiducial correspondences between the cloth and body meshes are predefined, which are used as control points for cloth deformation. The deformed cloth is matched to the outer surface of the occupancy map with a Laplacian regularization~\cite{sorkine2007rigid} (see Appendix). Three garment topologies for each cloth piece are used, i.e., tops: sleeveless shirts (3,763 vertices and 7,261 faces), T-shirts  (6,533 vertices, 13,074 faces), and long-sleeve shirts (8,269 vertices and 16,374 faces), and bottoms: short (3,975 vertices and 7,842 faces), medium (5,872 vertices and 11,618 faces), and long pants (11,238 vertices and 22,342 meshes), which are manually matched to each subject.

\vspace{-1mm}
\section{Evaluation}
\vspace{-1mm}
We evaluate HUMBI in terms of generalizability, diversity, and accuracy. For the generalizability, we conduct the cross-data evaluation on tasks of single view human reconstruction, e.g., monocular 3D face mesh prediction. For diversity, we visualize the distribution of HUMBI, e.g., gaze direction distribution along the yaw and pitch angle. For the accuracy, we measure how the number of cameras affects the quality of reconstruction. More evaluations can be found in Appendix.

% the gaze, headpose, and eye movement. 
% facehand mesh prediction from a single image. 
% accuracy (self consistency, e.g. symmetry, constant length), comparison with and augmentation to other dataset in terms of training networks, within-dataset and cross-dataset generalization capability.  

\begin{figure}[t]
\vspace{-3mm}
	\begin{center}
		\includegraphics[width=3.35in]{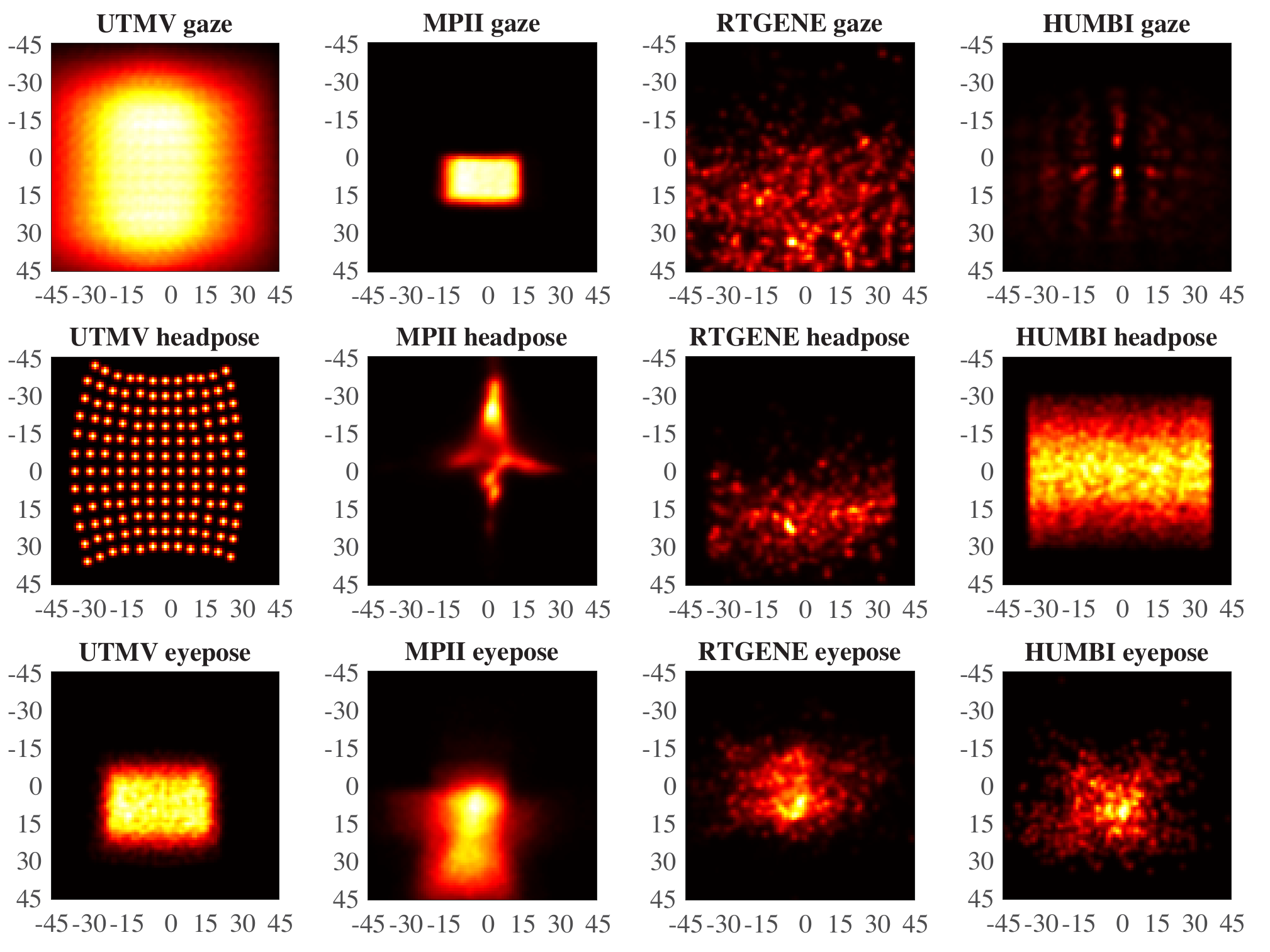}
	\end{center}
	\vspace{-7mm}
    \caption{Distribution of head pose, gaze and eye pose in normalized space for MPII-Gaze, UT-Multiview, RT-GENE and HUMBI. Horizontal and vertical axis represent yaw and pitch angle respectively (unit: degree).}
    \label{fig:gaze_distribution}
\end{figure}

{
\renewcommand{\tabcolsep}{4pt} % reduce default interval between text and column line
\begin{table}[t]
\vspace{3mm}
\setlength{\extrarowheight}{2pt}
\centering
% \footnotesize
\scriptsize
% \resizebox{0.995\linewidth}{!}
\begin{tabular}{l||c|c|c|c}
\hline
Bias / Variance  $\searrow$   &UTMV &MPII &RTGENE &HUMBI\\
\hline\hline
Gaze  &\textbf{7.43} / \textbf{33.09} &8.80 / 10.10 &19.35 / \underline{31.71} &\underline{7.70} / 30.01\\
\hline
Headpose &\underline{4.20} / \textbf{29.28} &12.51 / 16.04 &17.97 / 22.48 &\textbf{1.42} / \underline{24.77}\\
\hline
Eyepose &\underline{8.43} / 15.40 &20.81 / \underline{19.02} &\textbf{3.21} / 17.49 &8.78 / \textbf{19.04}\\
\hline
Average &\underline{6.69} / \textbf{25.93} &14.04 / 15.05 &13.51 / 23.90 &\textbf{5.98} / \underline{24.61}\\
\hline
\end{tabular}
\vspace{-3mm}
\caption{Bias and variance analysis of the distribution of head pose, gaze and eye pose (unit: degree, smallest bias and largest variance in bold, second with underline).}
\label{table:gaze_distribution_table}
\vspace{-3mm}
\end{table}
}

\subsection{Gaze}
\noindent\textbf{Benchmark Datasets} We use three benchmark datasets: (1) MPII-Gaze (MPII)~\cite{zhang15_cvpr} contains 213,659 images from 15 subjects, which was captured under the scenarios of everyday laptop use. (2) UT-Multiview (UTMV)~\cite{sugano:2014} is composed of 50 subjects with 160 gaze directions captured by 8 monitor-mounted cameras. Using the real data, the synthesized images from 144 virtual cameras are augmented. (3) RT-GENE~\cite{fischer2018rt} contains 122,531 images of 15 subjects captured by eye-tracking glasses.

\noindent\textbf{Distribution of Gaze Directions} To characterize HUMBI Gaze, we visualize three measures in Figure~\ref{fig:gaze_distribution}: (1) gaze pose: the gaze direction with respect to camera pose; (2) head pose: the head orientation with respect to the camera pose; and (3) eye pose: the gaze direction with respect to the head. HUMBI covers a wide and continuous range of head poses, due to numerous views and natural head movements by many subjects. The yaw and pitch of gaze and eye poses are distributed uniformly across all angles. The quantitative analysis of the bias and variance of the gaze distribution is summarized in Table~\ref{table:gaze_distribution_table}. HUMBI shows the smallest average bias (5.98$^\circ$ compared to 6.69$^\circ$-14.04$^\circ$ from other datasets) and second-largest average variance (24.61$^\circ$ compared to 25.93$^\circ$ of UTMV). Notice that UTMV is a synthesized dataset while HUMBI is real.

\noindent\textbf{Monocular 3D Gaze Prediction}
To validate the generalizability of HUMBI Gaze, we use an existing gaze detection network~\cite{zhang15_cvpr} to conduct a cross-data evaluation. We randomly choose $\sim$25K images (equally distributed among subjects) as experiment set for each dataset. One dataset is used for training and others are used for testing. Each data sample is defined as $\{(\mathbf{e}_{c},\mathbf{h}_{c}), \mathbf{g}_{c}\}$, where $\mathbf{e}_{c}\in\mathbb{R}^{36\times60}, \mathbf{h}_{c}\in\mathbb{R}^{2},\mathbf{g}_{c}\in\mathbb{R}^{2}$ are normalized eye patch, yaw and pitch angle of head pose, and gaze direction with respect to a virtual camera $c$. The detection network is trained to minimize the mean squared error of gaze yaw and pitch angles. We conduct a self-data evaluation for each dataset with 90\%/10\% of training/testing split. Table~\ref{table:gaze_experiment} summarize the experiment results. The detector trained by MPII and UTMV shows weak performance on cross-data evaluation comparing to HUMBI with 3$^\circ$-16$^\circ$ margin. HUMBI exhibits strong performance on cross-data evaluation with minimal degradation (less than 1$^\circ$ drop). Also, UTMV + HUMBI and MPII + HUMBI outperform each alone by a margin of 4.1$^\circ$ and 13.9$^\circ$ when tested on the third dataset MPII and UTMV respectively, showing that HUMBI is complementary to UTMV and MPII.
% because HUMBI covers a much wider range of head pose and diversity of eye appearance. 
% The qualitative results are shown in Figure~\ref{fig:}.

% \begin{table}[t]
% \setlength{\extrarowheight}{2pt}
% \centering
% \footnotesize
% \begin{tabular}{l|c|c|c}
% \hline
% % \multirow{3}{*}{Testing}  &  \multicolumn{3}{c}{Training}
% % \\[0.3ex]  
% % & 9s & 11s & 13s & 15s\\
% % \cline{2-4}
% \backslashbox[20mm]{Testing}{Training}    &  MPII-Gaze & UT-Multiview & HUMBI \\% & 9s & 11s & 13s & 15s\\
% \hline\hline
% MPII-Gaze  & 6.1$\pm$3.3 &11.8$\pm$6.6  &8.8$\pm$4.8 \\
% \hline
% UT-Multiview & 23.3$\pm$9.4 & 5.0$\pm$3.2 & 8.2$\pm$4.5  \\ %& 0.44$\pm$0.09 & 0.43$\pm$0.10 & \hline
% \hline
% HUMBI & 23.7$\pm$13.7 & 14.6$\pm$10.3 & 7.9$\pm$5.4 \\
% \hline
% \end{tabular}
% \vspace{-2mm}
% \caption{The mean error of 3D gaze prediction for the cross-data evaluation (unit: degree).}
% \label{table:gaze_experiment}
% \end{table}

% table for Gaze
{
\renewcommand{\tabcolsep}{3pt} % reduce default interval between text and column line
\begin{table}[t]
\setlength{\extrarowheight}{2pt}
\centering
% \footnotesize
\scriptsize
% \resizebox{0.995\linewidth}{!}
\begin{tabular}{l||c|c|c|c|c}
\hline
\multirow{2}{*}{\backslashbox[20mm]{Testing}{Training}} & \multirow{2}{*}{MPII} & \multirow{2}{*}{UTMV} & \multirow{2}{*}{HUMBI} & MPII & UTMV \\
& & & & + HUMBI & + HUMBI \\
\hline\hline
MPII  & 6.1$\pm$3.3 &11.8$\pm$6.6  &8.8$\pm$4.8 & 7.4$\pm$4.1 &7.7$\pm$4.6 \\
\hline
UTMV  & 23.3$\pm$9.4 & 5.0$\pm$3.2 & 8.2$\pm$4.5 &9.4$\pm$5.1 &5.4$\pm$3.2 \\
\hline
HUMBI & 23.7$\pm$13.7 & 14.6$\pm$10.3 & 7.9$\pm$5.4 & 8.9$\pm$6.2 &8.0$\pm$5.4 \\
\hline
\end{tabular}
\vspace{-2mm}
\caption{\small The mean error of 3D gaze prediction for the cross-data evaluation (unit: degree).}
\label{table:gaze_experiment}
\end{table}
}

% \begin{figure}[t]
% 	\begin{center}
% 		\includegraphics[width=3in]{./figure/exp_face_qual2.pdf}
% 	\end{center}
% 	\vspace{-5mm}
%     \caption{\small The qualitative results on 3D face mesh prediction trained with different dataset combination. The top and bottom show the results tested on 3DDFA and HUMBI dataset respectively.}
%     \label{fig:face_res}
% \end{figure}

% Recently this task is solved in a data-driven way, e.g., a deep neural network learns the statistic distribution of the dataset in inference time, as 3D geometry prediction from 2D appearance (image) is ill-posed problem,
\subsection{Face}
\noindent\textbf{Benchmark Dataset} We use 3DDFA~\cite{zhu2016face} that provides $\sim$6K 2D-3D pairs of the 3D face geometry and the associated images. We use 90\%/10\% of training/testing split. The base face model of 3DDFA is the Basel model~\cite{bfm09}, which is different from our face model (Surrey~\cite{huber2016multiresolution}). We manually pre-define the correspondences between two models in the canonical coordinates. 

\begin{figure}[t]
	\begin{center}
\hspace{-4mm}\includegraphics[width=0.5\textwidth]{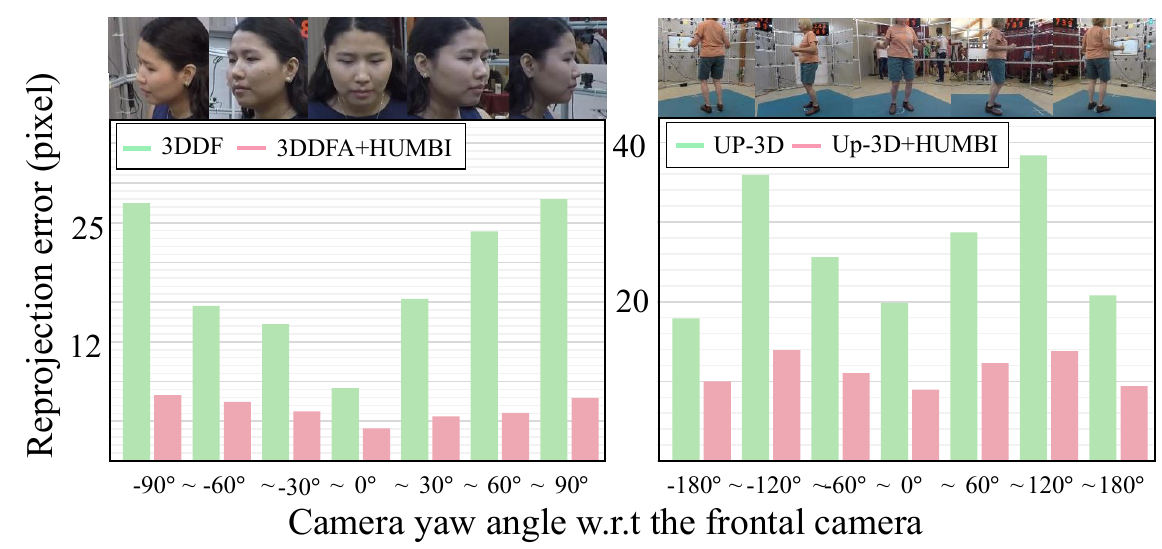}
	\end{center}
\vspace{-5mm}
    \caption{We measure viewpoint dependency of a face/body mesh reconstruction model trained by multiple datasets. Augmenting HUMBI substantially reduce the view dependency.}
    \label{fig:view_aug}
\end{figure}

%  As our base face model (Surrey~\cite{huber2016multiresolution}) does not match to the one of 3DDFA (Basel~\cite{bfm09}), we pre-define the correspondences between the two different faces in canonical space and then match the vertex indices.
\noindent\textbf{Monocular 3D Face Mesh Prediction}\label{exp:monoface} 
We evaluate HUMBI Face by predicting a 3D face mesh using a recent mesh reconstruction network~\cite{Yoon_2019_CVPR}. The network encoder directly regresses the 3D face shape and head pose from a single view image. We modify the decoder to accommodate the differentiable Basel model. We train the network with three dataset combinations, i.e., 3DDFA, HUMBI, and 3DDFA+HUMBI, and for each training, we minimize the loss of the reprojection error with weak perspective projection model. To measure the accuracy, we use the reprojection error scaled to the input image resolution (256 pixel). Table~\ref{table:face_experiment} summarize the results. From the results of 3DDFA+HUMBI, the prediction accuracy is improved from both datasets (2.8 pixels from 3DDFA and 4.9 pixels from HUMBI) by combining two datasets, which indicates the complementary nature of HUMBI. Due to the multiview images in HUMBI, the network can learn the view-invariant geometric representation, which allows precise reconstruction even with considerable occlusion as shown in Figure~\ref{fig:view_aug}.

\begin{table}[t]
\setlength{\extrarowheight}{2pt}
\centering
\footnotesize
\begin{tabular}{l|c|c|c}
\hline
% \multirow{3}{*}{Testing}  &  \multicolumn{3}{c}{Training}
% \\[0.3ex]  
% & 9s & 11s & 13s & 15s\\
% \cline{2-4}
\backslashbox[20mm]{Testing}{Training}    &  3DDFA & HUMBI & 3DDFA+HUMBI \\% & 9s & 11s & 13s & 15s\\
\hline\hline
3DDFA & 7.1$\pm$6.4 & 20.7$\pm$7.1  &\textbf{4.3$\pm$6.6} \\
\hline
HUMBI & 23.5$\pm$13.9 & 13.3$\pm$13.7 & \textbf{8.4$\pm$12.2}  \\
\hline
\end{tabular}
\vspace{-2mm}
\caption{The mean error of 3D face mesh prediction for cross-data evaluation (unit: pixel).}
\label{table:face_experiment}
\end{table}

% \begin{table}[t]
% \setlength{\extrarowheight}{2pt}
% \centering
% \scriptsize
% \caption{The cross-dataset validation results on the monocular 3D face mesh prediction task. The mean and standard deviation of the reprojection error is measured. Unit: pixel}
% \begin{tabular}{l|c|c|c}
% \hline
% \multirow{2}{*}{Testing dataset}  &  \multicolumn{3}{c}{Training dataset (face mesh)}
% \\[0.3ex]  
% % & 9s & 11s & 13s & 15s\\
% \cline{2-4}

%     &  3DDFA & HUMBI & 3DDFA$+$HUMBI \\% & 9s & 11s & 13s & 15s\\
% \hline
% 3DDFA & 7.168$\pm$6.4 & 20.736$\pm$7.168  &\textbf{4.352$\pm$6.656} \\
% \hline
% HUMBI & 23.552$\pm$19.968 & 13.312$\pm$19.712 & \textbf{8.422$\pm$19.2}  \\ %& 0.44$\pm$0.09 & 0.43$\pm$0.10 & \hline
% \hline
% \end{tabular}
% \label{table:face_experiment}
% \end{table}

% that provides ~ 3D face reconstruction with the associated 2D image pairs

\subsection{Hand}
\noindent\textbf{Benchmark Datasets} We use three benchmark datasets: (1) Rendered Handpose Dataset (RHD)~\cite{zimmermann2017learning} is a synthesized hand dataset containing 44K images built from 20 freely available 3D models performing 39 actions.  (2) Stereo Hand Pose Tracking Benchmark (SHPTB)~\cite{zhang2017hand} is a real hand dataset captured by a stereo rgb camera rig. (3) FreiHAND~\cite{Freihand2019} is a multi-view real hand dataset captured by 8 cameras. (4) ObMan~\cite{hasson19_obman} is a large scale synthetic hand mesh dataset with associated 2D images (141K pairs). We use previous two datasets for the hand keypoint evaluation and the last one for the hand mesh evaluation.

\noindent\textbf{Monocular 3D Hand Pose Prediction}
To validate HUMBI Hand, we conduct a cross-data evaluation for the task of the 3D hand pose estimation from a single view image, where we use a recent hand pose detector~\cite{zimmermann2017learning}.
% to predict 3D hand pose from a set of 2D hand score maps
% \zhixuan{Zhixuan: can I remove "from a set of 2D hand score maps" here in order to solve R1's confusion mentioned in detailed comments?}. 
We train and evaluate the model trained by each dataset and a combination of HUMBI and each other dataset. The results are summarized in Table~\ref{table:handpose_experiment}. We use area under PCK curve (AUC) in an error range of 0-20mm as the metric. It show that HUMBI is more generalizable for predicting 3D hand pose than other three dataset (by a margin of 0.02-0.16 AUC). Moreover, HUMBI is complementary to other datasets and the performance of model trained by another dataset alone is increased with HUMBI (by a margin of 0.04-0.12 AUC).

% We compare the generalizability of two networks trained on HUMBI (116K hand instances from 100 subjects) and RHD (41K instances), respectively, by predicting on SHPTB (6k images). We use three metrics for experiment: average mean end-point error (EPE), average median end-point error and area under PCK curve (AUC). Results show that the HUMBI is highly generalizable for predicting 3D hand pose of the third dataset (7\% higher detection rate than RHD) in Table~\ref{table:handpose_experiment}. 

% table for hand keypoint
% \begin{table}[t]
% \setlength{\extrarowheight}{2pt}
% \centering
% \footnotesize
% \begin{tabular}{l|c|c|c}
% \hline
% \backslashbox[20mm]{Dataset}{Metric} & EPE mean & EPE median & AUC \\
% \hline\hline
% RHD & 32.302 mm & 31.322 mm & 0.400 \\
% \hline
% HUMBI & 27.977 mm & 26.889 mm & 0.473  \\
% \hline
% \end{tabular}
% \vspace{-2mm}
% \caption{Experiment result of network trained by RHD and HUMBI while test on SHPTB dataset. AUC of PCK is calculated over an error range of 0-20 mm.}
% \label{table:handpose_experiment}
% \end{table}

\begin{table}[t]
\setlength{\extrarowheight}{2pt}
\centering
% \footnotesize
\scriptsize
% \resizebox{0.995\linewidth}{!}
\begin{tabular}{l||c|c|c|c|c|c|c}
\hline
\backslashbox[20mm]{Testing}{Training}&S &R &F &H &S+H &R+H &F+H\\
\hline\hline
STB (S)  &0.72 &0.40 &0.22 &0.47 &0.40 &0.52 &0.44\\
\hline
RHD (R) &0.16 &0.59 &0.26 &0.49 &0.48 &0.50 &0.44 \\
\hline
FreiHand (F) &0.15 &0.40 &0.72 &0.37 &0.35 &0.43 &0.35 \\
\hline
HUMBI (H) &0.16 &0.36 &0.18 &0.50 &0.43 &0.47 &0.41 \\
\hline\hline
Average &0.30 &0.44 &0.36 &\textbf{0.46} &\textbf{0.42} &\textbf{0.48} &\textbf{0.41} \\
\hline
\end{tabular}
\vspace{-2mm}
\caption{Cross-data evaluation results of 3D hand keypoint prediction. Metric is AUC of PCK calculated over an error range of 0-20 mm.}
\vspace{-3mm}
\label{table:handpose_experiment}
\end{table}

% to RHD by providing a numerous views.
% Notice that it is evaluated in 3D space over all 21 hand keypoints.

% in terms of training a pose prior network that estimating the most likely 3D structure conditioned on the score maps of 2D hand pose\cite{zimmermann2017learning}, testing on Stereo Hand Pose Tracking Benchmark~\cite{zhang20163d}. Both datasets provide 2D and 3D annotations of 21 hand keypoints, which is consistent with ours. In our experiment, a network is trained on 116K hand samples from 100 subjects of HUMBI while another same network is trained on all 41K samples of RHD. Both networks are then evaluated on one subject (6K images) in Stereo Hand Pose Tracking Benchmark dataset. We use Percentage of Correct Keypoints (PCK) as the metric and experiment results is shown in ~\ref{fig:hand_pose_pck}. Notice that it is evaluated in 3D space over all 21 hand keypoints.

% \begin{figure}[t]
% 	\begin{center}
% 		\includegraphics[clip,width=3.3in]{./figure/hand_pck.png}
% 	\end{center}
% 	\label{fig:hand_pose_pck}
% 	\vspace{-5mm}
%     \caption{\small PCK curve showing the performance of the network describe in [??] trained by RHD and HUMBI, evaluated on Stereo Hand Pose Tracking[??] dataset}
% \end{figure}

\noindent\textbf{Monocular 3D Hand Mesh Prediction}
We compare HUMBI Hand with synthetic ObMan~\cite{hasson19_obman} dataset. We use a recent regression network~\cite{Yoon_2019_CVPR} that outputs the hand mesh shape and camera pose with minor modifications, e.g., we change the size of the latent coefficient and the hand mesh decoder to the ones from the MANO hand model. We train and evaluate the network based on the reprojection error with weak perspective projection model. The results are summarized in Table~\ref{table:hand_experiment}. Due to the domain gap between the real and synthetic data, the prediction accuracy of the network trained with synthetic data is largely degraded on the real data. However, by combining two datasets, the performance is highly improved (even better than intra-data evaluation), e.g., ObMan+HUMBI can outperform ObMan and HUMBI 0.3 and 1.7 pixels, respectively.

\subsection{Body}
\noindent\textbf{Benchmark Datasets} We use four benchmark datasets: (1) Human3.6M~\cite{h36m_pami} contains numerous 3D human poses of 11 actors/actresses measured by motion capture system with corresponding images from 4 cameras. (2) MPI-INF-3DHP~\cite{mehta2017monocular} is 3D human pose estimation dataset, which contains both 3D and 2D pose labels as well as images covering both indoor and outdoor scenes. We use its test set containing 2,929 valid frames from 6 subjects. 
% (3) MPII dataset is a large scale in-the-wild 2D human pose dataset containing 25K training images (it was only used to augment ). 
(3) UP-3D~\cite{lassner2017unite} is a 3D body mesh dataset providing $\sim$9K pairs of 3D body reconstruction and the associated 2D images. We use Human3.6M, MPI-INF-3DHP for body pose evaluation and UP-3D for body mesh evaluation.

\noindent\textbf{Monocular 3D Body Pose Prediction}
To validate HUMBI body, we conduct a cross-data evaluation for the task of estimating 3D human pose from a single view image. We use a recent body pose detector~\cite{zhou2017towards}.
% We compared the generalizability of two networks with same architecture trained on two combinations: MPII (23K instances) + HUMBI (23K instances from 100 subjects) and MPII (23K instances) + Human3.6M (23K instances), respectively, by predicting on MPI-INF-3DHP (3K instances). Experiment shows that the network trained with HUMBI outperforms Human3.6M for commonly defined keypoints, especially on lower body parts, with an overall $\sim$4\% margin (Table~\ref{table:body_experiment}).
%We use Percentage of Correct Keypoints (PCK) as the metric and experiment results is shown in ~\ref{fig:body_pose_pck}. Notice that it is evaluated in 3D space over 14 body joints(1 for neck, 1 for pelvis, 3 for each leg, 3 for each arm) that are defined in a common way among all datasets involved.
We train and evaluate model trained by each dataset and model trained by a combination of HUMBI and each other dataset. By following the training protocol of~\cite{zhou2017towards}, we use 2D landmark labels from MPII dataset~\cite{andriluka14cvpr} for a weak supervision. The results are summarized in Table~\ref{table:body_experiment}. We use area under PCK curve (AUC) in an error range of 0-150 mm as the metric. It show that HUMBI is more generalizable for predicting 3D body pose than Human3.6M and MPI-INF-3DHP (by a margin of 0.023 and 0.064 AUC). Moreover, HUMBI is complementary to each other dataset and the performance of model trained by another dataset alone is increased with HUMBI (by a margin of 0.057 and 0.078 AUC respectively).

\begin{table}[t]
\setlength{\extrarowheight}{2pt}
\centering
\footnotesize
\begin{tabular}{l|c|c|c}
\hline
% \multirow{3}{*}{Testing}  &  \multicolumn{3}{c}{Training}
% \\[0.3ex]  
% & 9s & 11s & 13s & 15s\\
% \cline{2-4}
\backslashbox[20mm]{Testing}{Training}    &  ObMan & HUMBI & ObMan+HUMBI \\% & 9s & 11s & 13s & 15s\\
\hline\hline
ObMan & 3.84$\pm$2.6 & 6.1$\pm$4.1 &\textbf{3.5$\pm$2.4} \\
\hline
HUMBI & 10.6$\pm$11.3 & 6.5$\pm$8.4 & \textbf{4.8$\pm$5.8}  \\ %& 0.44$\pm$0.09 & 0.43$\pm$0.10 & \hline
\hline
\end{tabular}
\vspace{-2mm}
\caption{The mean error of 3D hand mesh prediction for cross-data evaluation (unit: pixel).}
\label{table:hand_experiment}
\vspace{-3mm}
\end{table}

\begin{table}[t]
\setlength{\extrarowheight}{2pt}
\centering
% \footnotesize
\scriptsize
% \resizebox{0.995\linewidth}{!}
\begin{tabular}{l||c|c|c|c|c}
\hline
\multirow{2}{*}{\backslashbox[20mm]{Testing}{Training}} 
& \multirow{2}{*}{H36M} & \multirow{2}{*}{MI3D} & \multirow{2}{*}{HUMBI} & H36M &MI3D\\
& & & & +HUMBI & +HUMBI \\
\hline\hline
H36M  &0.562 &0.362 &0.434 &0.551 &0.437\\
\hline
MI3D &0.317 &0.377 &0.354 &0.375 &0.425 \\ %& 0.44$\pm$0.09 & 0.43$\pm$0.10 & \hline
\hline
HUMBI &0.248 &0.267 &0.409 &0.372 &0.377 \\
\hline\hline
Average &0.376 &0.335 &\textbf{0.399} &\textbf{0.433} &\textbf{0.413} \\
\hline
\end{tabular}
\vspace{-2mm}
\caption{Cross-data evaluation results of 3D body keypoint prediction. Metric is AUC of PCK calculated over an error range of 0-150 mm.}
\vspace{-3mm}
\label{table:body_experiment}
\end{table}

\begin{table}[t]
\setlength{\extrarowheight}{2pt}
\centering
\footnotesize
\begin{tabular}{l|c|c|c}
\hline
% \multirow{3}{*}{Testing}  &  \multicolumn{3}{c}{Training}
% \\[0.3ex]  
% & 9s & 11s & 13s & 15s\\
% \cline{2-4}
\backslashbox[20mm]{Testing}{Training}    &  UP-3D & HUMBI & UP-3D+HUMBI \\% & 9s & 11s & 13s & 15s\\
\hline\hline
UP-3D & 22.7$\pm$18.6 &49.4$\pm$0.09 &\textbf{18.4$\pm$13.8} \\
\hline
HUMBI & 26.0$\pm$19.7 & 14.5$\pm$6.6 & \textbf{12.5$\pm$8.4}  \\ %& 0.44$\pm$0.09 & 0.43$\pm$0.10 & \hline
\hline
\end{tabular}
\vspace{-2mm}
\caption{The mean error of 3D body mesh prediction for cross-data evaluation (unit: pixel).}
\vspace{-2mm}
\label{table:body_experiment1}
\end{table}

\noindent\textbf{Monocular 3D Body Mesh Prediction}
We compare the body mesh prediction accuracy using a recent CNN model trained on (1) HUMBI, (2) UP-3D, and (3) HUMBI+UP-3D. While we use~\cite{Yoon_2019_CVPR} for the testing CNN model, recent monocular body reconstruction methods~\cite{alldieck2019tex2shape, alldieck2018video, alldieck2018detailed, alldieck19cvpr, bhatnagar2019mgn, pifuSHNMKL19, habermann2019livecap, lazova2019360,SMPL-X:2019, hmrKanazawa17, omran2018neural, rong2019delving} can be alternative to test the generalization ability of HUMBI. The network decoder is modified to accommodate the differentiable SMPL parameter prediction. The reprojection error is used to supervise the network and to evaluate testing performance. The cross-data evaluation is summarized in Table~\ref{table:body_experiment1}. We observe that the network trained with HUMBI shows weak performance because of the lack of diversity of poses. However, it is highly complementary to other datasets as it provides various appearance from 107 viewpoints as shown in Figure~\ref{fig:view_aug}.

\begin{figure}[t]
% 	\begin{center}
% 		\includegraphics[clip,width=3.3in]{./figure/camera_ablation.pdf}
% 	\end{center}
% 	\vspace{-5mm}
%     \caption{\small Camera ablation order. Cameras are uniformly removed during the ablation test in Figure~\ref{fig:camera_ablation_study}. }
%     \label{fig:ablation}
%     \vspace{-6mm}
	\begin{center}
	\vspace{-2mm}
		\includegraphics[width=0.5\textwidth]{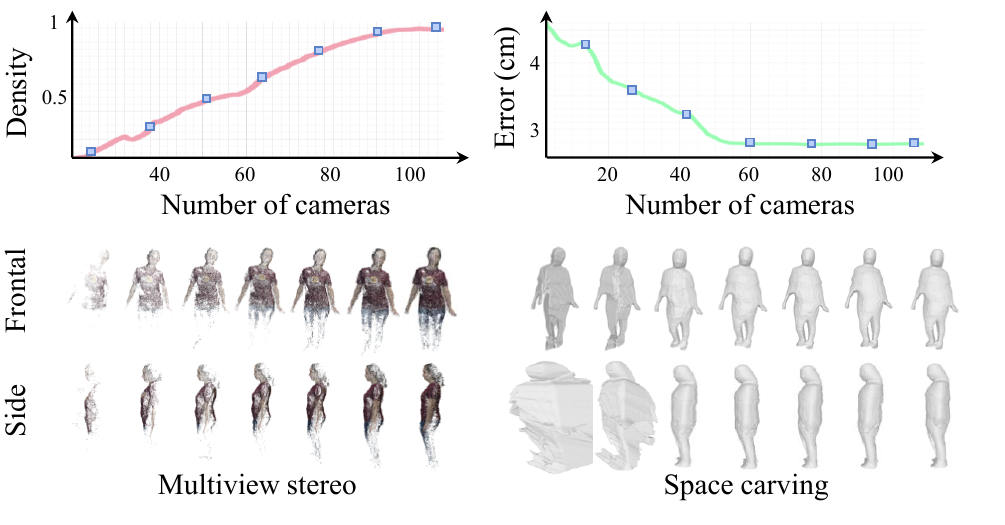}
	\end{center}
    \label{fig:ablation}
	\vspace{-7mm}
    \caption{We conduct camera-ablation study to evaluate the accuracy of the garment reconstruction in terms of the density (left) and the accuracy (right).}
    \label{fig:camera_ablation_study}
\end{figure}

\subsection{Garment}
We conduct camera-ablation study to evaluate how the number of cameras affect garment reconstruction quality. We incrementally reduce the number of cameras and measure the reconstruction accuracy and density. The reconstruction density is computed by the number of 3D points produced by multiview stereo~\cite{schoenberger2016sfm}. The reconstruction accuracy metric is the closest point distance from the 3D garment surface reconstructed by shape-from-silhouette~\cite{Laurentini94}. In both cases, the performance reaches to the optimal even without 107 cameras as shown in Figure~\ref{fig:camera_ablation_study}, ensuring that our garment reconstruction is accurate (density: 90 cameras $\approx$ 107 cameras; accuracy: 60 cameras $\approx$ 107 cameras). The additional evaluations on the garment silhouette accuracy can be found in the Appendix.

\section{Discussion}
We present HUMBI dataset that is designed to facilitate high resolution pose- and view-specific appearance of human body expressions. Five elementary body expressions (gaze, face, hand, body, and garment) are captured by a dense camera array composed of 107 synchronized cameras. The dataset includes diverse activities of 772 distinctive subjects across gender, ethnicity, age, and physical condition. We use a 3D mesh model to represent the expressions where the view-dependent appearance is coordinated by its canonical atlas. Our evaluation shows that HUMBI outperforms existing datasets as modeling nearly exhaustive views and can be complementary to such datasets. 

HUMBI is the first-of-its-kind dataset that attempts to span the general appearance of assorted people by pushing towards two extremes: views and subjects. This will provide a new opportunity to build a versatile model that generates photorealistic rendering for authentic telepresence. However, the impact of HUMBI will not be limited to appearance modeling, i.e., it can offer a novel multiview benchmark dataset for a stronger and generalizable reconstruction and recognition model specific to humans. 

\section*{Acknowledgement}
This work was partially supported by National Science Foundation (No.1846031 and 1919965), National Research Foundation of Korea, and Ministry of Science and ICT of Korea (No. 2020R1C1C1015260).

% This material is based upon work supported by the National Science Foundation under Grants No.1846031 and 1919965.
% This work was partially supported by the National Research Foundation of Korea (NRF) grant and funded by the Korea government (MSIT) (No. 2020R1C1C1015260).

% to push the boundary of human modeling and rendering. 

% 107 multi-camera system

% capture natural human behavioral signals using a 107 multi-camera system. As partly shown in Fig.~\ref{fig:teaser_big}, \ref{fig:multi-view}, \ref{fig:face}, \ref{fig:additional_results}, this dataset includes 164 distinctive subjects across gender, ethnicity, age, physical condition where we provide 3D high fidelity computational models for gaze, face, finger, body, and cloth. We believe that this dataset will take us to the next level of human behavioral understanding, which will make a signifcant impact on millions of people's daily lives. 

\balance
{\small
\bibliographystyle{ieee}
\bibliography{egbib}
}

%%%%%%%%%%%%%%%%%%%%%%%%%%%%%%%%%%%%%%%%%%%%%%%%%%%%%%%%%%%%%%%%%%%%%
%%%%%%%%%%%%%%%%%%%%%%%%%%%%%%%%%%%%%%%%%%%%%%%%%%%%%%%%%%%%%%%%%%%%%
\clearpage
\begin{appendices}

% commented out following part for arXiv
% \setcounter{section}{}
% \def\thesection{\Alph{section}}
% \renewcommand{\thesubsection}{\thesection.\arabic{subsection}}
% \maketitle

This supplementary material provides additional details of HUMBI.

\section{Multi-camera System}\label{sec:system}

We design a unique multi-camera system that was deployed in public events including Minnesota State Fair and James Ford Bell Museum of Natural History at the University of Minnesota. There are 772 subjects captured by 107 GoPro HD cameras recording at 60Hz.

\noindent\textbf{Hardware} The capture stage is made of a re-configurable dodecagon frame with 3.5 m diameter and 2.5 m height using T-slot structural framing (80/20 Inc.). The stage is encircled by 107 GoPro HD cameras (38 HERO 5 BLACK Edition and 69 HERO 3+ Silver Edition), one LED display for an instructional video, eight LED displays for video synchronization, and additional lightings. Among 107 cameras, 69 cameras are uniformly placed along the two levels of the dodecagon arc (0.8 m and 1.6 m) for body and cloth, and 38 cameras are place over the frontal hemisphere for face and gaze.

\noindent\textbf{Performance Instructional Video}
To guide the movements of the participants, we create four instructional videos ($\sim$2.5 minutes). Each video is composed of four sessions. (1) Gaze: a subject is asked to find and look at the requested number tag posted on the camera stage; (2) Face: the subject is asked to follow 20 distinctive dynamic facial expressions (e.g., eye rolling, frowning, and jaw opening); (3) Hand: the subject is asked to follow a series of American sign languages (e.g., counting one to ten, greeting, and daily used words);  (4) Body and garment: the subject is asked to follow range of motion, which allows them to move their full body and to follow slow and full speed dance performances curated by a professional choreographer.
% a subject in capture system watches few instructional videos, and he/she is asked to follow or perform a requested movement. 
% Four video clips displaying 2 minutes for each; (1) find and look at the requested number tag posted on the camera stage, (2) follow 20 distinctive dynamic facial expressions (e.g., eye rolling, frowning, and jaw opening), (3) follow range of motion, which allows them to move their full body, (4) follow a dance performed by a professional choreographer.

\noindent\textbf{Synchronization and Calibration} We manually synchronize 107 cameras using LED displays. The maximum synchronization error is up to 15 ms. We use the COLMAP~\cite{schoenberger2016sfm} software for camera synchronization, and upgrade the reconstruction to metric scale by using the physical distance between cameras and the ground plane.

\section{HUMBI Reconstruction}\label{sec:recon}
Given the synchronized multiview image streams, we reconstruct body expressions in 3D. 

\subsection{3D Keypoint Reconstruction}\label{sec:keypoint}
Given a set of synchronized and undistorted multiview images, we detect 2D keypoints of face, hand, body (including feet)~\cite{cao2017realtime}. 
% A geometric verification process based on certain pivot body keypoints is performed across views to group detected 2D keypoints according to the real person they belongs to. 
Using these keypoints, we triangulate 3D keypoints with RANSAC~\cite{Fischler:1981} followed by the non-linear refinement by minimizing reprojection error~\cite{hartley:2004}\footnote{When multiple persons are detected, we use a geometric verification to identify each subject.}. 

\begin{figure}[h]
	\begin{center}
		\includegraphics[width=0.45\textwidth]{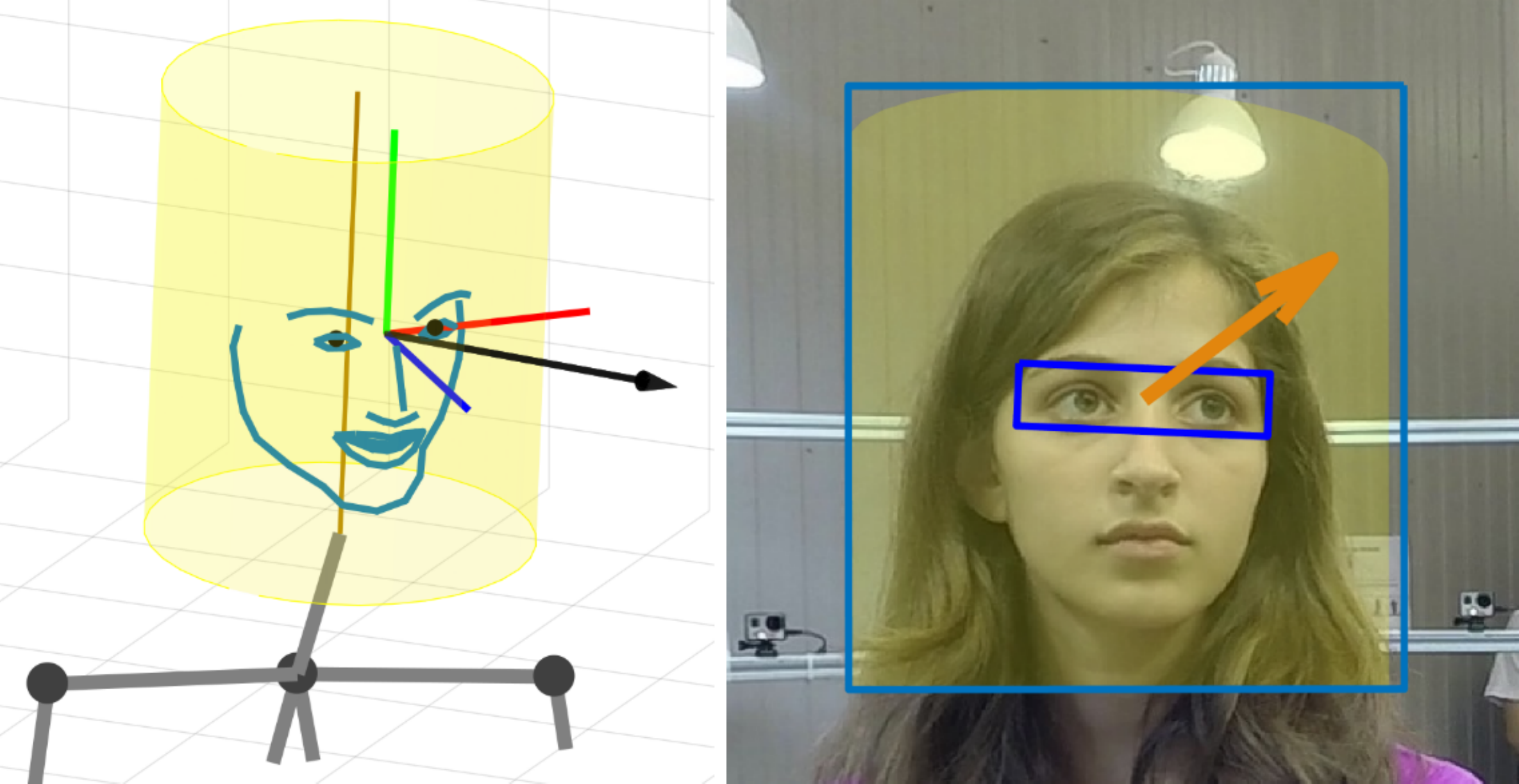}
	\end{center}
	\vspace{-5mm}
    \caption{\small Gaze signals computed by our system (Sec.~\ref{sec:gaze}). (Left) 3D demonstration of captured gaze placed on the black dotted body joints. Black arrow is gaze direction. Red, green and blue segment are $x$, $y$ and $z$-axis of gaze frame. Brown segment is the center axis of the head cylinder. (Right) Gaze overlaid on a color image. Orange arrow is gaze direction. Dark blue box indicates eye region. Blue box wraps face. Yellow area is projection of the cylinder.}
    \label{fig:gaze3}
\end{figure}

In the RANSAC process, we apply the length constraint (e.g., symmetry between left and right body) and reason about visibility of keypoints based on confidence of detection, camera proximity, and viewing angle. 

\subsection{Gaze}\label{sec:gaze}

We define the moving coordinate of gaze using facial keypoints. Figure~\ref{fig:gaze3} illustrates the moving coordinate. The black arrow is gaze direction. The red, green and blue segments are $x$, $y$ and $z$-axis of gaze frame. The brown segment is the center axis of the head cylinder. On the right, the orange arrow is the gaze direction. Dark blue box indicates eye region. Blue box wraps face. Yellow area is projection of the cylinder.

%%% face
\begin{figure}[t]
	\begin{center}
	    \includegraphics[width=0.45\textwidth]{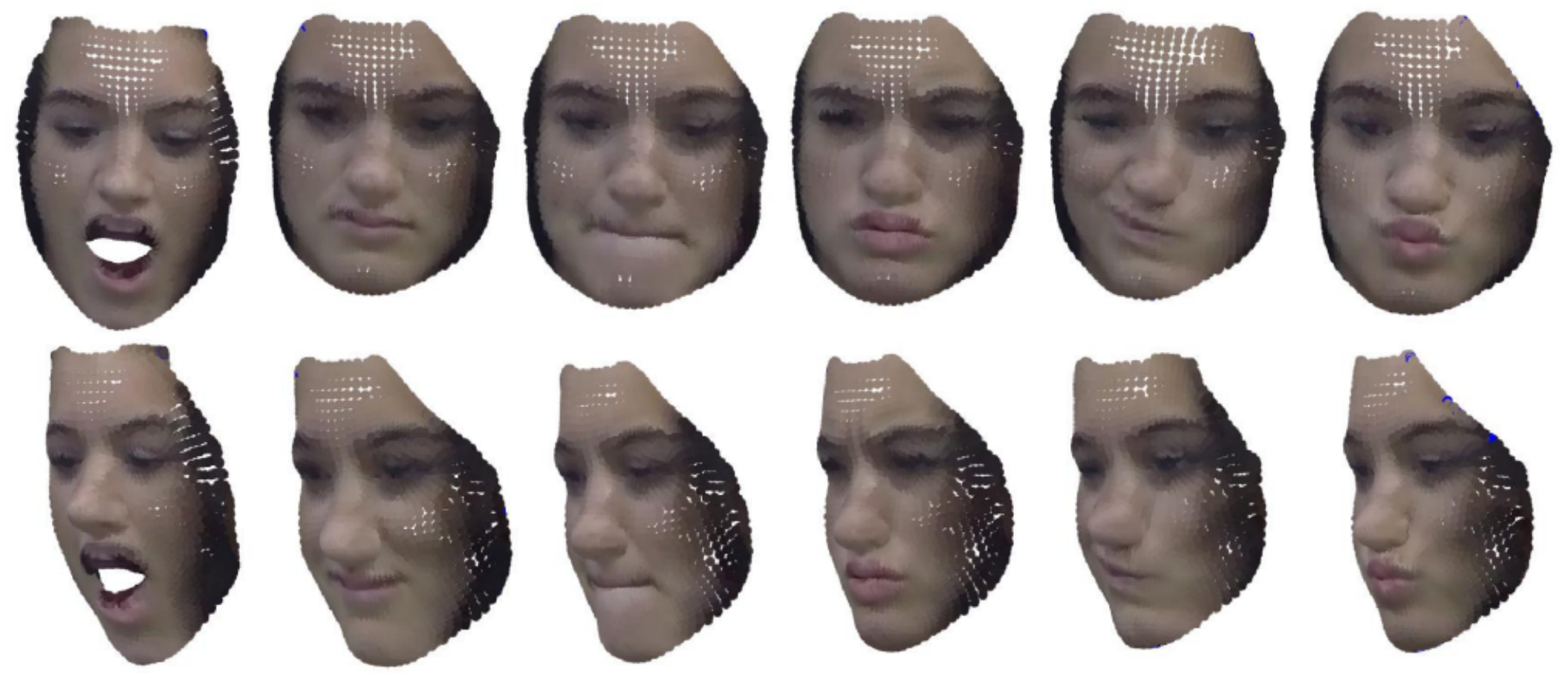}
		\includegraphics[width=0.45\textwidth]{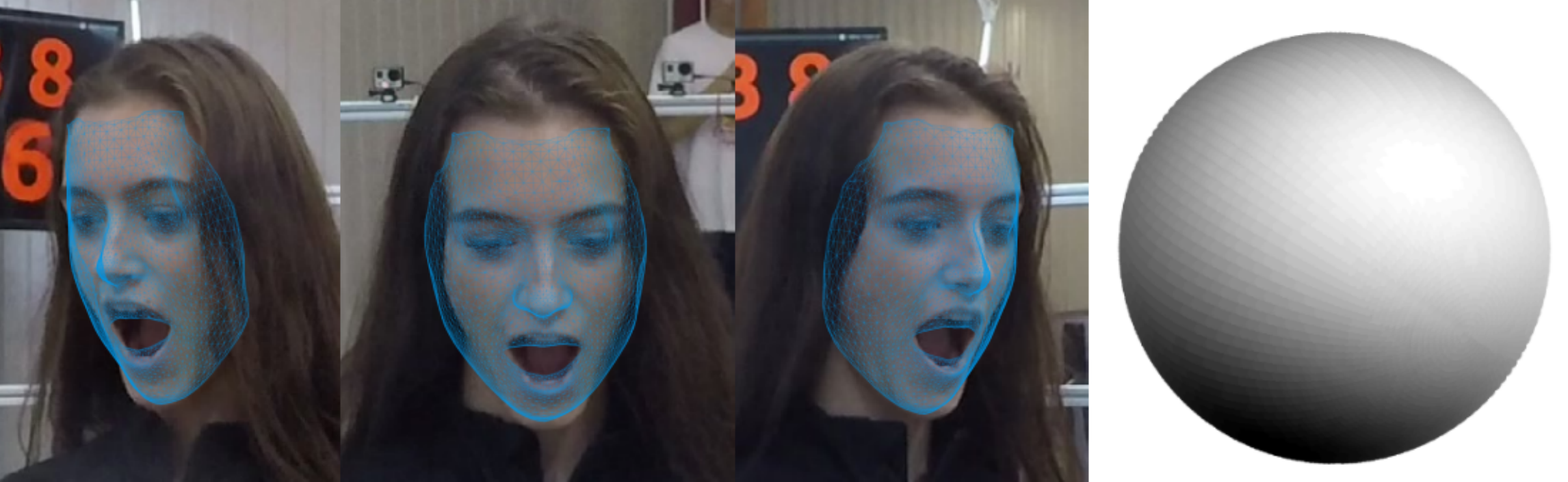}
	\end{center}
	\vspace{-5mm}
    \caption{\small Face reconstruction (Section~\ref{sec:face}). (Top) Recovered 3D faces with various expressions (Bottom left) Alignment between projected mesh and subject's face. (Bottom right) Estimated illumination condition.}
    \label{fig:face}
\end{figure}

\subsection{Face}\label{sec:face}

We model $\mathcal{M}_{\rm face} = f_{\rm face}(\mathcal{K}_{\rm face}, \mathcal{I}_{\rm face})$. We represent a face mesh using Surrey face model\cite{huber2016multiresolution}, which is a 3D morphable model (3DMM) defined as: 
\begin{align}
    \mathcal{V}_{\rm face}(\boldsymbol{\alpha}^s,\boldsymbol{\alpha}^e) = \mathbf{S}_0 + \sum_{i=1}^{K_s} \alpha_i^s \mathbf{S}_i + \sum_{i=1}^{K_e} \alpha_i^e \mathbf{E}_i,
\end{align}
where $\mathcal{V}_{\rm face}\in \mathbb{R}^{3D_s}$ is the 3D face vertices, $\mathbf{S}_0$ is the meanface, $\mathbf{S}_i$ and $\alpha_i^s$ are the $i^{\rm th}$ shape basis and its coefficient, and $\mathbf{E}_i$ and $\alpha_i^e$ are the $i^{\rm th}$ expression basis and its coefficient. $D_s$ is the number of points in the shape model. 
% The shape parameters capture subject identity, and the expression parameters describe its vary with different facial expressions. 
% In practice, we only use first 10 shape basis for robustness.

The model is fitted to multiview images $\mathcal{I}_{\rm face}$ by minimizing the following cost:
\begin{align}
    E_{\rm face} = E^{k}_{\rm face} + \lambda_{face}^{a} E^{a}_{\rm face}, \label{eq:face}
\end{align}
where $E^{k}_{\rm face}$ and $E^{a}_{\rm face}$ are errors of 3D keypoint and appearance, respectively.

% \vspace{2mm}\noindent\textbf{Shape and Expression}

We minimize the geometric error between 3D face model and the reconstructed keypoints:
\begin{align}
E^{k}_{\rm face}\big(\mathbf{Q},\boldsymbol{\alpha}^s,\boldsymbol{\alpha}^e) = \sum_i^{68} \|\mathcal{K}_{\rm face}^i- \mathbf{Q}(\overline{\mathbf{V}}_{\rm face}^i)\|^2 \nonumber
\end{align}
where $\boldsymbol{\alpha}^s\in\mathbb{R}^{63}$ and $\boldsymbol{\alpha}^e\in\mathbb{R}^{6}$ are shape and expression coefficients, $\mathcal{K}_{\rm face}^i$ is $i^{\rm th}$ face keypoint, and $\overline{\mathbf{V}}_{\rm face}^i$ is the corresponding $i^{\rm th}$ vertex in $\mathbf{V}_{\rm face}$. $\mathbf{Q}$ is a 6D rigid transformation between the 3DMM in its cannonical coordinate system and the reconstructed model in the world coordinate system.

% \vspace{2mm}\noindent\textbf{Appearance} 
For appearance fitting, we use text model from Basel Face Model\cite{bfm09}:
\begin{align}
    \mathbf{T} = \mathbf{T}_0 + \sum_{i=1}^{K_t} \alpha_i^t \mathbf{T}_i,
\end{align}
where $\mathbf{T}\in \mathds{R}^{3\times D_s}$ is the 3D face texture, $\mathbf{T}_0$ is the mean texture model, $\mathbf{T}_i$ and $\alpha_i^t$ are the $i^{\rm th}$ texture basis and its coefficient.

The appearance model is combination of texture and illumination: $\mathbf{C} = \mathbf{I}(\mathcal{V}_{\rm face}, \mathbf{T}, \boldsymbol{\alpha}^h)$ where $\mathbf{C}$ is the RGB color for a 3D face and $\mathbf{I}$ uses Lambertian illumination to estimate the appearance. We model the illumination using the spherical harmonics basis model where $\boldsymbol{\alpha}^h$ is the coefficient for the harmonics. From this, the error of appearance is:
\begin{align}
    E^{a}_{\rm face}(\boldsymbol{\alpha}^s, \boldsymbol{\alpha}^e, \boldsymbol{\alpha}^t, \boldsymbol{\alpha}^h) =  \sum_j \|\mathbf{c}_j - \phi_j (\mathbf{C})\|^2,
\end{align}where $\phi_j (\mathbf{C})$ is the projection of the appearance $\mathbf{C}$ onto the $j^{\rm th}$ camera, and $\mathbf{c}_j$ is the face appearance in the $j^{\rm th}$ image.

We optimize Equation~(\ref{eq:face}) using a nonlinear least squares solver with ambient light initialization. Figure~\ref{fig:face} illustrate the resulting face reconstruction where we compute the shape, expression, texture and reflectance. To learn the consistent shape of the face model for each subject, we infer the maximum likelihood estimate of the shape parameter given the reconstructed keypoints over frames, which allows us to fit to the best model (Figure~\ref{fig:face}).

%%% hand
\subsection{Hand}\label{sec:hand}

We model $\mathcal{M}_{\rm hand}(\boldsymbol{\theta}_h, \boldsymbol{\beta}_h)=f_{\rm hand}(\mathcal{K}_{\rm face})$. We represent a hand mesh using the MANO parametric hand model~\cite{MANO:SIGGRAPHASIA:2017}, which is composed of 48 pose parameters and 20 shape parameters
% defined as 
% \begin{align}
% \mathcal{V}_{\rm hand} = f_{\rm mano}(\mathbf{H};\boldsymbol{\theta}_h, \boldsymbol{\beta}_h), 
% \end{align}
% where $\mathbf{H}$ is hand vertices taken from SMPL full body model as template, 
where $\boldsymbol{\theta}$ and $\boldsymbol{\beta}$ are the pose and shape parameters, respectively. 

We minimize the following objective to model $f_{\rm hand}$:
\begin{align}
E_{\rm hand}\big(\boldsymbol{\theta},\boldsymbol{\beta}) = E_{\rm hand}^k + \lambda_{h}^\theta E_{\rm hand}^\theta + \lambda_{h}^\beta E_{\rm hand}^\beta,
\end{align}
where $\lambda_\theta$ and $\lambda_\beta$ are weights for pose and shape regularization, respectively.

Given the correspondence between the reconstructed keypoints and the hand mesh, we minimize their error:
\begin{align}
    E_{\rm hand}^k(\boldsymbol{\theta},\boldsymbol{\beta}) = \sum_i \|\mathcal{K}^i_{\rm hand} - \mathbf{Q} (\overline{\mathcal{V}}_{\rm hand}^i)\|^2,
\end{align}
where $\mathcal{Q}$ is the rigid transformation between the keypoints and the hand mesh model in its canonical coordinate system. 

We apply regularization on shape and pose parameters:
\begin{align}
    E_{\rm hand}^\theta (\boldsymbol{\theta},\boldsymbol{\beta}) = \|\boldsymbol{\theta}\|^2, E_{\rm hand}^\beta = \|\boldsymbol{\beta}\|^2.
\end{align}
Rigid transformation parameters are firstly estimated by aligning 6 keypoints on palm, then shape and expression parameters are estimated alternatively until converge, followed by nonlinear optimization for all parameters. For the same subject, initially hand mesh of each frame is reconstructed independently. Then shape parameters are fixed as the median values of all frames. Other parameters are optimized, subsequently.

%%% body
\begin{figure}[b]
	\begin{center}
		\includegraphics[width=0.45\textwidth]{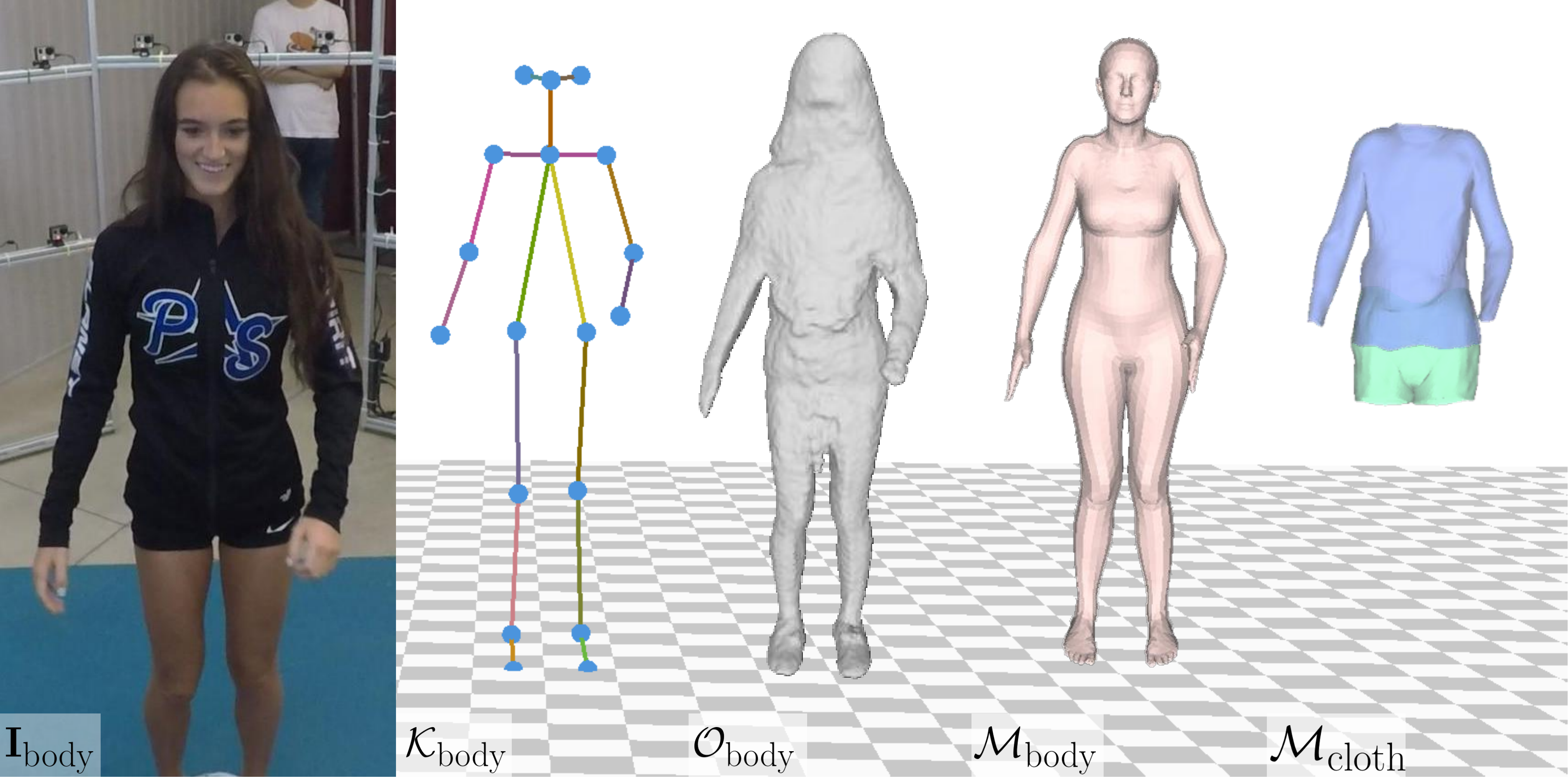}
	\end{center}
	\vspace{-4mm}
    \caption{\small HUMBI body and cloth reconstruction results.}
    \label{fig:smpl}
\end{figure}

% \in\mathds{R}^{4129}
% \in\mathds{R}^{7999}
% \in\mathds{R}^{4129}
% \in\mathds{R}^{7999}
\subsection{Body}\label{sec:body}
We model $\mathcal{M}_{\rm body}=f_{\rm body}(\mathcal{K}_{\rm body}, \mathcal{O}_{\rm body})$. We represent the body expression using a parametric SMPL model~\cite{loper2015smpl} and fit to the 3D body keypoints $\mathcal{K}_{\rm body}$ and the occupancy map $\mathcal{O}_{\rm body}$ by minimizing the following objective: 
\begin{align}
E_{\rm body}(\boldsymbol{\alpha}_{b}, \boldsymbol{\beta}_{b}, \boldsymbol{\theta}_{b})= E^{p}_{\rm body}+ \lambda^{s}_{b}E^{s}_{\rm body}+ \lambda^{r}_{b}E^{r}_{\rm body},
\label{eq:optim_body}
\end{align}
where $\lambda^{s}_{b}$ and $\lambda^{r}_{b}$ control the importance of each measurement. $\boldsymbol{\beta}_{\rm b}\in \mathbb{R}^{10}$ represents the linear shape coefficient, and $\boldsymbol{\alpha}_{\rm b}\in \mathbb{R}^{72}$ represents Euler angles for the 24 joints (one root joint and 23 relative joints between body parts). $\boldsymbol{\theta}_{\rm body}\in\mathds{R}^{4}$ denotes the translation and scale of the mean body. 

We prescribe the correspondence between the pose of SMPL model with 3D body keypoints, i.e., $\overline{\mathcal{V}}_{\rm body}^i$ is the $i^{\rm th}$ keypoint of the SMPL. $E^{p}_{\rm body}$ penalizes the distance between the reconstructed 3D body keypoints $\mathcal{K}_{\rm body}$ and the keypoints of the SMPL $\overline{\mathcal{V}}_{\rm body}$:
\begin{align}
E^{p}_{\rm body}(\boldsymbol{\alpha}_{b},\ \boldsymbol{\theta}_{b})=\sum_i \left\|\mathcal{K}_{\rm body}^i - \overline{\mathcal{V}}_{\rm body}^{i}\right\|^2.
\label{eq:data}
\end{align}

$E^{s}_{\rm body}$ encourages the shape of the estimated body model $\mathcal{M}_{\rm body}$ to be aligned with the outer surface of the occupancy map $\mathcal{O}_{\rm body}$. We use Chamfer distance to measure the alignment: %at every optimization solver:
\begin{align}
 &E^{s}_{\rm body}(\boldsymbol{\alpha}_{b},\ \boldsymbol{\beta}_{b},\ \boldsymbol{\theta}_{b}) =  d_{\rm chamfer}(\mathcal{O},\mathcal{V}_{\rm body}),
\label{eq:icp2}
\end{align}
where $d_{\rm chamfer}$ measures Chamfer distance between two sets of point clouds.

% $\tau$ represents ICP function.
% &\mathrm{where}~~
%  \tau(\mathcal{O},\mathcal{M}_{\rm body})  = \underset{i}{\operatorname{argmin}}~~ \|\mathcal{O}-\mathcal{M}^{i}_{\rm body}\|^2,

$E^r_{body}$ penalizes the difference between the estimated shape $\boldsymbol{\beta}_{b}$ and the subject-aware mean shape $\boldsymbol{\beta}^{\rm prior}_{b}$ as follows:
 \begin{align}
 E^{r}_{\rm body}(\boldsymbol{\beta}_{b};\boldsymbol{\beta}^{\rm prior}_{b})= \left\|\boldsymbol{\beta}_{b} - \boldsymbol{\beta}^{\rm prior}_{b}\right\|^2.
\label{eq:prior}
\end{align}
This prevents unrealistic shape fitting due to the estimation noise/error, e.g., long hair covering body.
To obtain the shape prior $\boldsymbol{\beta}^{prior}_{b}$, we solve the Eq.~(\ref{eq:optim_body}) without $E_{r}^{\rm body}$ using the recovered volumes of the same subject and take the median $\boldsymbol{\beta}_{b}$ for robustness. 

\begin{figure}[b]
	\begin{center}
\hspace{-4mm}\includegraphics[clip,width=3.4in]{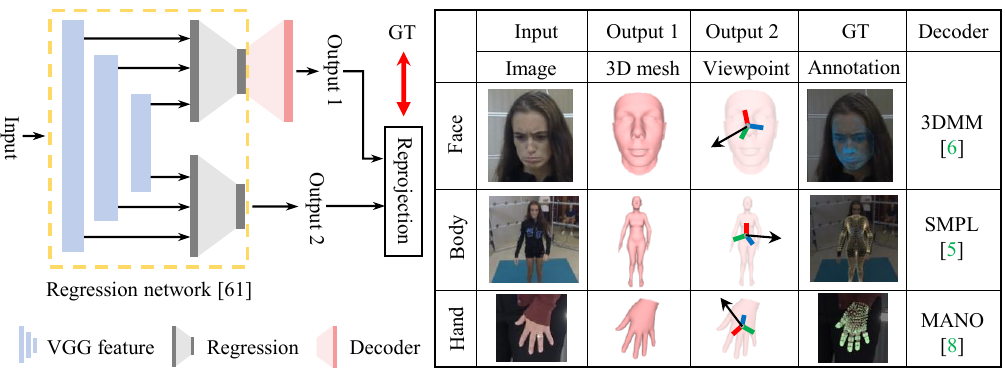}
	\end{center}
\vspace{-4mm}
    \caption{\small {The training setup for 3D mesh prediction from a single image.}}
    \label{fig:network}
\end{figure}

\begin{figure}[t]
	\begin{center}
		\includegraphics[width=0.45\textwidth]{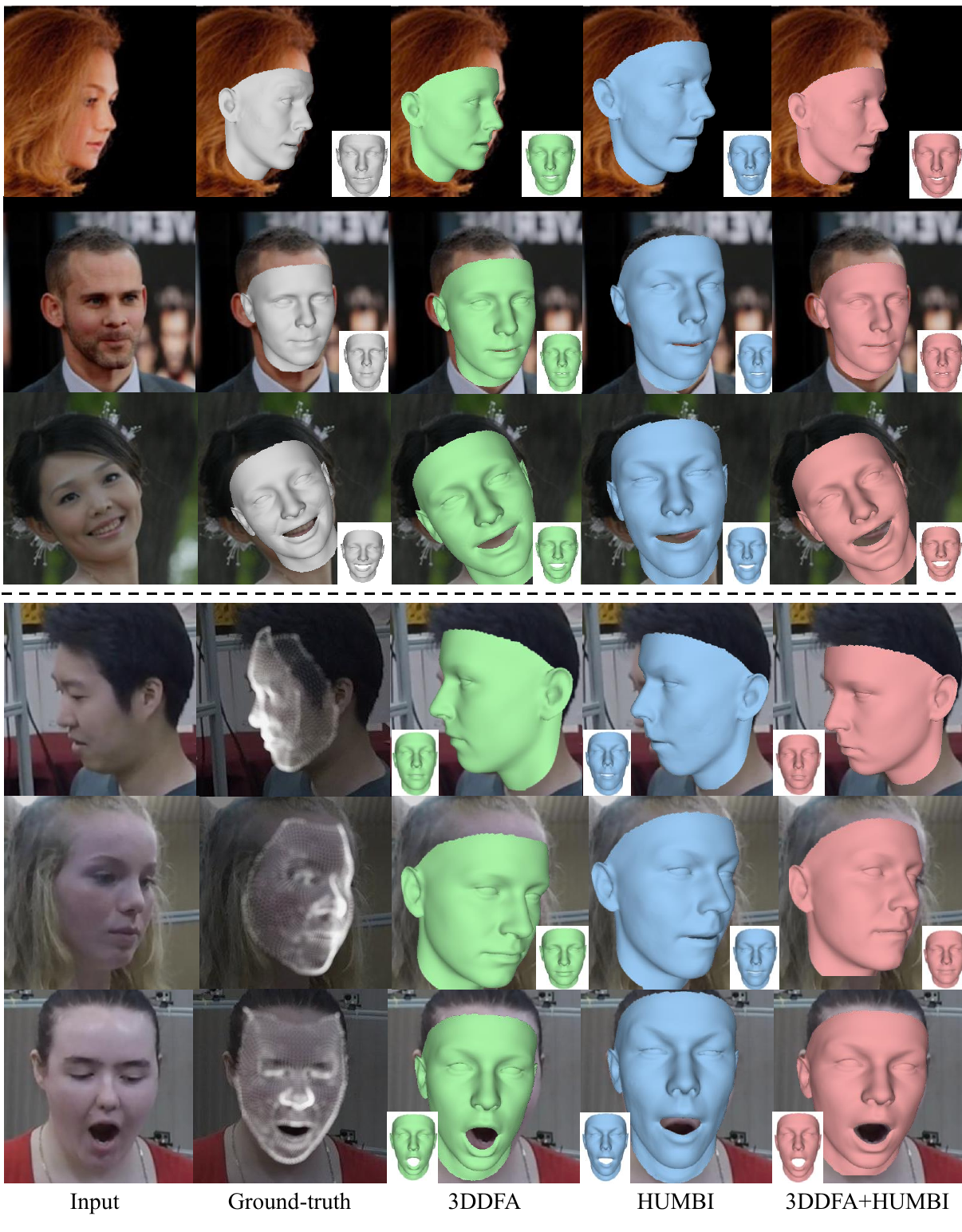}
	\end{center}
	\vspace{-5mm}
    \caption{\small The qualitative results of the monocular 3D face prediction network trained with different dataset combination. The top and bottom show the testing on the external and HUMBI Face respectively.}
    \label{fig:face_res}
\end{figure}

\subsection{Garment}\label{sec:cloth}
We model a garment fitting function $\mathcal{M}_{\rm cloth}=f_{\rm cloth}(\mathcal{M}_{\rm body},\mathcal{O}_{\rm body})$ by representing the garment with an in-house mesh model $\mathcal{M}_{\rm cloth}$. The assumption of the minimally clothed body shape~\cite{pons2017clothcap} is made. We minimize the following objective:
% to fit the garment to the 3D body $\mathcal{M}_{\rm body}$ and the outer surface of the occupancy map $\mathcal{O}_{\rm body}$:
\begin{align}
E_{\rm cloth}(\mathbf{R}_{c},\mathbf{t}_{c})=E_{\rm cloth}^{b}+\lambda_{c}^{o}E_{\rm cloth}^{o}+\lambda_{c}^{r}E_{\rm cloth}^{r},
\label{eq:optim}
\end{align}
where $\lambda_{c}^{o}$ and $\lambda_{c}^{r}$ control the importance of each measurement. 

We manually establish the set of correspondences between $\mathcal{M}_{\rm body}$ and $\mathcal{M}_{\rm cloth}$ that move approximately the same way. $E_{\rm cloth}^{b}$ measures the correspondence error:
\begin{align}
E_{\rm cloth}^{b}(\mathcal{V}_{\rm cloth})=\sum_{i}\|\overline{\mathcal{V}}_{\rm body}^{i}-\overline{\mathcal{V}}_{\rm cloth}^{i}\|^2,
\end{align}
where $\overline{\mathcal{V}}_{\rm body}$ and $\overline{\mathcal{V}}_{\rm cloth}$ are the corresponding vertices.

$E_{\rm cloth}^{o}$ measures the Chamfer distance to align  $\mathcal{M}_{\rm cloth}$ with $\mathcal{O}_{\rm body}$:
\begin{align}
    E_{\rm cloth}^{o}(\mathcal{V}_{\rm cloth}) =  d_{\rm chamfer}(\mathcal{O}_{\rm body},\mathcal{V}_{\rm cloth}).
\end{align}

$E_{\rm cloth}^{r}$ is the spatial regularization (Laplacian) that prevents from reconstructing unrealistic cloth structure by penalizing a non-smooth and non-rigid vertex with respect to its neighboring vertices~\cite{sorkine2007rigid}:
\begin{align}
    E_{\rm cloth}^{r} = \nabla^2 \mathcal{M}_{\rm cloth}.
\end{align}

\begin{figure}[t]
	\begin{center}
		\includegraphics[width=0.5\textwidth]{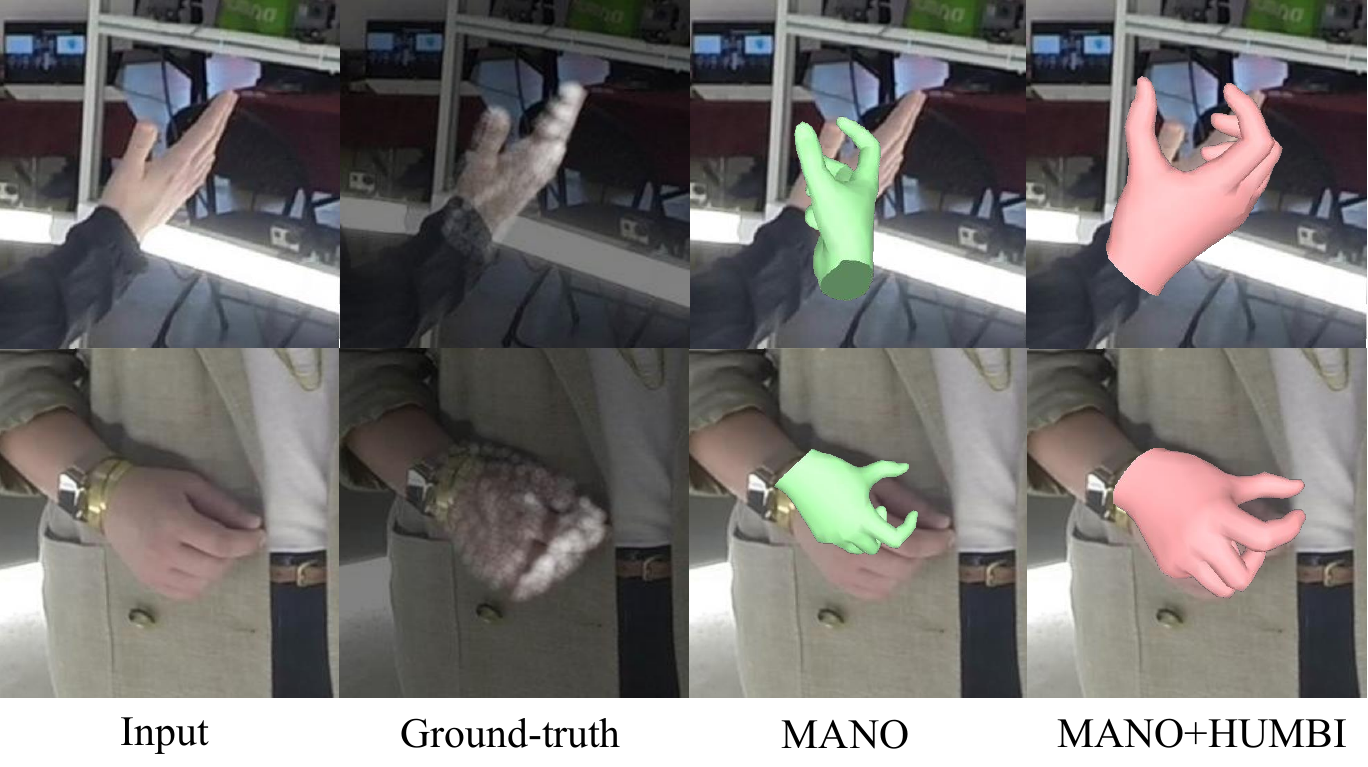}
	\end{center}
	\vspace{-5mm}
    \caption{\small Monocular 3D hand mesh prediction results tested on HUMBI Hand.}
    \label{fig:hand_res}
\end{figure}

\begin{figure}[t]
	\begin{center}
		\includegraphics[width=0.5\textwidth]{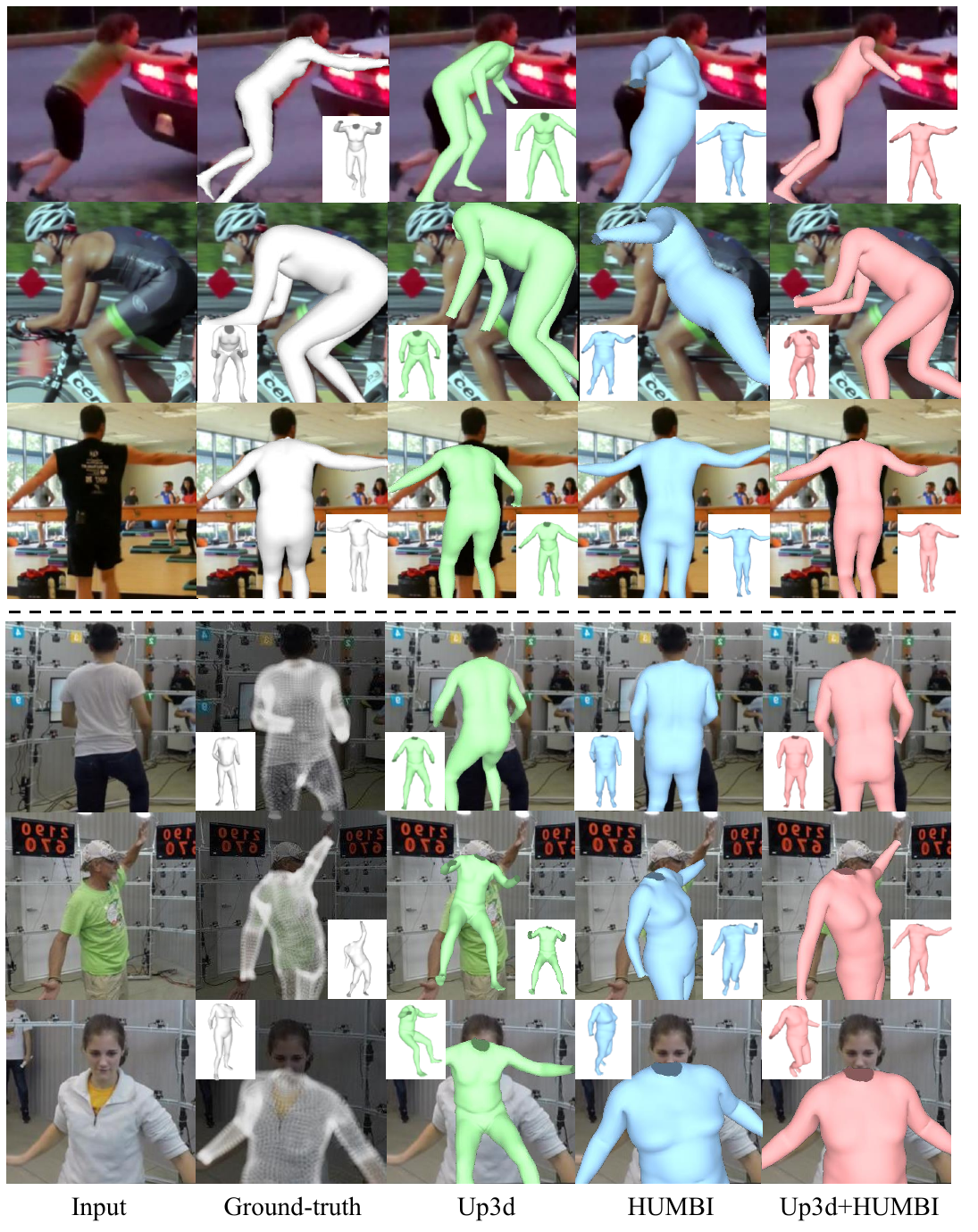}
	\end{center}
	\vspace{-5mm}
    \caption{\small The qualitative results of the monocular 3D body prediction network trained with different dataset combination. The top and bottom show the results tested on UP-3D and HUMBI Body respectively.}
    \label{fig:smpl2}
\end{figure}

\section{Training Mesh Prediction Network}\label{sec:mesh_train}
To train the mesh prediction function of each body expression (i.e., face, hand, and body described in Section 4.1-4.3 of the main paper), we use the recent neural network~\cite{Yoon_2019_CVPR} that can regress a single image to the body model parameters, e.g., SMPL body shape and pose coefficients, and camera viewpoint. In Figure~\ref{fig:network}, the encoder is implemented with \cite{Yoon_2019_CVPR}, and the decoder with the pre-trained weights of each body model, i.e., 3DMM~\cite{bfm09} for face, SMPL~\cite{loper2015smpl} for body, and MANO~\cite{MANO:SIGGRAPHASIA:2017} for hand. The network is trained by minimizing the reprojection error where only the regression network is newly trained. The training details are described in Figure~\ref{fig:network}.

\section{More Results}\label{sec:qual_res}
\subsection{Mesh Prediction Results}\label{sec:mesh_pred}
We use a recent CNN model to evaluate HUMBI as introduced in Section~\ref{sec:mesh_train}. The qualitative evaluation on single view prediction is shown in Figure~\ref{fig:face_res} (face), Figure~\ref{fig:hand_res} (hand), and Figure~\ref{fig:smpl2} (body).

\subsection{Garment Reconstruction Accuracy}\label{sec:garment}
\begin{figure}[h]
	\begin{center}
	\vspace{-3mm}
		\includegraphics[width=3in]{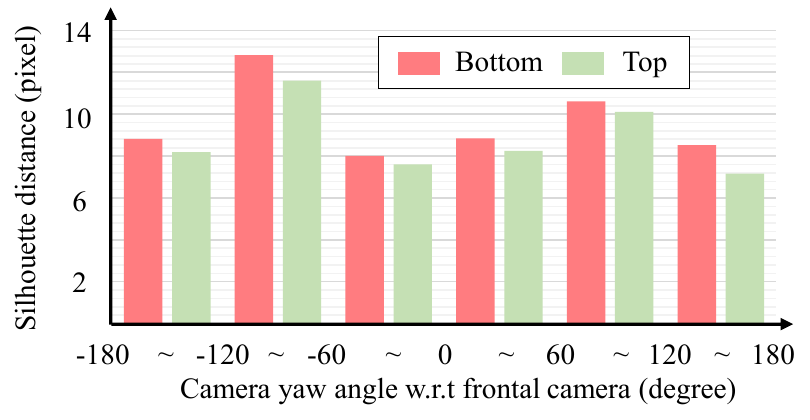}
	\end{center}
	\vspace{-7mm}
    \caption{\small Garment silhouette error.}
    \label{fig:garment}
	\vspace{-5mm}
\end{figure}

We provide additional evaluation of view-dependent garment silhouette accuracy measured by the Chamfer distance between the annotated and the reprojected garment boundary in 2D. We pick a half-sleeve shirts and half pants models as a representative garment of top and bottom and measure the accuracy from each camera view that has different angle with respect to the most frontal camera. On average in Figure~\ref{fig:garment}, the silhouette error seen from the side view (11 pixels) is higher than the frontal (7.5 pixels) and rear views (8 pixels). 

\end{appendices}

\end{document}